\documentclass[pre,preprint]{revtex4-1}

\usepackage[left=1in,right=1in,top=1in,bottom=1in]{geometry}
\usepackage{graphicx}
\usepackage{epstopdf}
\usepackage{amsmath}
\usepackage{mathtools}
\usepackage{amssymb}
\usepackage{upgreek}
\usepackage{float}
\usepackage[table]{xcolor}
\usepackage[normalem]{ulem}
\graphicspath{{./figures/},{./figures_eps/}}
\usepackage{contour}
\usepackage{subcaption}
\usepackage{tikz}

\usepackage{verbatim}
\usepackage{ulem}
\usepackage{color,soul}
\usepackage{enumerate}

\usepackage{setspace}

\newcommand{\MDGrevise}[1]{\textcolor{black}{#1}}
\newcommand{\KZrevise}[1]{\textcolor{black}{#1}}
\newcommand{\KZr}[1]{\textcolor{black}{#1}}

\newcommand{\eff}{\textnormal{eff}}

\newcommand{\xt}{\tilde{u}}
\newcommand{\reals}{\mathbb{R}}
\newcommand{\rank}{\mathrm{rank}}
\newcommand{\loss}{\mathcal{L}}
\newcommand{\trace}{\mathrm{tr}}
\newcommand{\Irnn}{I^{rd_ud_u}}
\newcommand{\Irnm}{I^{rd_ud_z}}
\newcommand{\Irmn}{I^{rd_zd_u}}
\newcommand{\Irmm}{I^{rd_zd_z}}
\newcommand{\Ehat}{\hat{E}}
\newcommand{\Lhat}{\hat{W}}
\newcommand{\Dhat}{\hat{D}}
\newcommand{\Ecal}{\mathcal{E}}
\newcommand{\Lcal}{\mathcal{L}}
\newcommand{\Wcal}{\mathcal{W}}
\newcommand{\Dcal}{\mathcal{D}}
\newcommand{\Ccal}{\mathcal{C}}

\usepackage{subcaption}
\usepackage[bookmarksnumbered=true,breaklinks=true,colorlinks,citecolor=blue,linkcolor=blue]{hyperref}

\begin{document}

\title{
Autoencoders for discovering manifold dimension and coordinates in data from complex dynamical systems 
}

\author{Kevin Zeng}
\author{Carlos E.~P\'erez De Jes\'us}
\author{Andrew J.~Fox}
\author{Michael D. Graham}
\email{Email: mdgraham@wisc.edu}
\affiliation{Department of Chemical and Biological Engineering, University of Wisconsin-Madison, Madison WI 53706, USA}
\date{\today}

\begin{abstract}
While many phenomena in physics and engineering are formally high-dimensional, their long-time dynamics often live on a lower-dimensional manifold. The present work introduces an autoencoder framework that combines implicit regularization with internal linear layers and $L_2$ regularization (weight decay) to automatically estimate the underlying dimensionality of a data set, produce an  orthogonal manifold coordinate system, and provide the mapping functions between the ambient space and manifold space, allowing for out-of-sample projections. We validate our framework's ability to estimate the manifold dimension for a series of datasets from dynamical systems of varying complexities and compare to other state-of-the-art estimators. We analyze the training dynamics of the network to glean insight into the mechanism of low-rank learning and find that collectively each of the implicit regularizing layers compound the low-rank representation and even self-correct during training. Analysis of gradient descent dynamics for this architecture in the linear case reveals the role of the internal linear layers in leading to faster decay of a ``collective weight variable" incorporating all layers, and the role of weight decay in breaking degeneracies and thus driving convergence along directions in which no decay would occur in its absence. We show that this framework can be naturally extended for applications of state-space modeling and forecasting by generating a data-driven dynamic model of a spatiotemporally chaotic partial differential equation using only the manifold coordinates. Finally, we demonstrate that our framework is  robust to hyperparameter choices.
\end{abstract}

\maketitle
\section{Introduction} \label{Introduction}

\newcommand{\IM}{\mathcal{M}}
Nonlinear dissipative partial differential equations (PDEs) are ubiquitous in describing phenomena throughout physics and engineering that display complex nonlinear behaviors, out-of-equilibrium dynamics, and even spatiotemporal chaos. Although {the state space of a PDE is formally infinite-dimensional}, the long-time dynamics of a  dissipative system are known or suspected to collapse onto a finite-dimensional invariant manifold, which we will denote $\IM$.  \citep{Hopf:1948bn,Teman1977,Zelik.2022.10.48550/arxiv.2208.12101}. {The same idea holds for high-dimensional dissipative systems of ordinary differential equations (or discretized PDEs), and in any case, data from any system under consideration will be finite-dimensional, so we will consider manifolds of dimension $d_m$}  embedded in an ambient space $\mathbb{R}^{d_u}$, where often  $d_m\ll d_u$. That is to say, in order to accurately describe the manifold, and thus the underlying dynamics of the system, only $d_m$ independent coordinates are needed (at least locally). {In general, no global coordinate representation of dimension $d_m$ is available, but Whitney's theorem guarantees that a global representation with \emph{embedding dimension}  
$d_e\leq 2d_m$ can be found \cite{Lee:2012bn}. Alternately, in principle, an atlas of overlapping charts with dimension $d_m$ can be constructed to provide local $d_m$-dimensional representations \citep{Lee:2012bn,FloryanCANDyMan2022}.} \KZrevise{For the most part, we address the task of learning minimal \emph{global} manifold representations (although we will show that our work can be extended into local representations)}{, and consider cases where $d_e=d_m$.}


Obtaining a minimal manifold coordinate description for these systems {based on an analysis of data from that system} is ideal for a number of dynamical applications such as state-space identification, reduced-order modeling {and control}, and system interpretability, as well as many other downstream tasks such as classification. However, estimating the underlying dimensionality {of a data set} and obtaining the manifold coordinate transformations is generally a nontrivial task. Given access only to {data represented in} the high-dimensional ambient space of a system, the challenge becomes the following: 1) determining $d_m$, 2) constructing a coordinate system describing points in $\mathcal{M}$, and 3) obtaining the mapping functions $\mathcal{E}:\mathbb{R}^{d_u} \rightarrow\mathbb{R}^{d_m}$ and $\mathcal{D}:\mathbb{R}^{d_m}\rightarrow \mathbb{R}^{d_u}$. In the interest of identifying and modeling the underlying core dynamics of these systems, our aim is to address these three challenges  using a single framework trained on high-dimensional ambient data alone.


These three challenges have been tackled by an extensive variety of methodologies, but we emphasize that rarely are all three challenges addressed simultaneously in a single framework --often only the first challenge of identifying the manifold dimension is attempted. For complex dynamical systems, many of these methods developed in systems theory rely on high-precision analyses and access to the underlying equations. For example, \citet{Yang2009,Yang2012} estimated the manifold dimension of the Kuramoto-Sivashinsky equation (KSE), a formally infinite-dimensional system with finite-dimensional dynamics, for a range of parameters using covariant Lyapunov vectors, monitoring when the Lyapunov spectrum of the system begins to rapidly fall. \citet{Ding2016} corroborated these results, estimating the dimension of \KZr{the} invariant manifold containing the long time dynamics of the KSE for a domain size of $L=22$ via a Floquet mode approach applied to organized unstable periodic orbits identified in the system.  These methods require high precision \KZrevise{solutions of the governing equations and access to very specific dynamical data (e.g.\ periodic orbits)} that for more complex systems such as the Navier-Stokes equations are nontrivial or even intractable tasks. Furthermore, these methods are not applicable when the governing equations are not known or when data is collected {from general time series rather than precisely prescribed trajectories.} For these reasons, these methods will not be the focus of this work.

Towards more generalized and data-driven approaches, the task of estimating the number of degrees of freedom required to represent a set of data without loss of information has been explored in the fields of pattern recognition, information sciences, and machine learning. These methods produce estimates of $d_m$ (or some upper bound) either using global or local analyses of the dataset.

Global approaches tackle this challenge in several ways. Linear global projection methods, such as Principal Component Analysis (PCA) \cite{Jolliffe1986} and its variants (e.g. Sparse PCA \cite{Zou2004} and Bayesian PCA \cite{Bishop1998}), {determine a linear} subspace in which the projection of the data minimizes some projection error. These methods are useful in that not only are they computationally tractable, they also directly provide the mapping functions to the low dimensional representation. However, as they are linear methods, they generally overestimate $d_m$, {since representing data on a curved manifold of dimension $d_m$ will require at least $d_m+1$ coordinates.} 

Nonlinear PCA, or deep autoencoders in general, deal with nonlinearity using neural networks tasked with autoassociation \cite{KARHUNEN1994}. Autoencoders can be used to estimate $d_m$ by tracking the mean squared reconstruction error (MSE) as a function of the bottleneck dimension $d_z$ of the networks. If the MSE significantly drops above a threshold value of $d_h$, one can infer that the minimum number of degrees of freedom needed to represent the system data is reached. In applications toward complex high-dimensional dynamical systems including discretized dissipative PDEs, \citet{Linot2020,Linot2021} and \citet{Vlachas2022} used undercomplete autoencoders to estimate the manifold dimension of data from the KSE this way. However, as system complexity and dimensionality increase, the MSE drop off becomes less and less sharp \citep{Linot2020,Lipson2022,Jesus.2023.10.1103/physrevfluids.8.044402,Linot.2023.10.1017/jfm.2023.720}. Additionally, a practical drawback of this type of approach is it requires training separate networks with a range of $d_z$.

Towards more automated autoencoder-based frameworks, several works have incorporated the heuristic false-nearest neighbor algorithm (FNN) \cite{Kennel1992} to target the embedding dimension for state-space reconstructions of a time-series signal that come from \KZrevise{systems with manifolds with $d_m=3$}. Specifically, \citet{Gilpin2020} incorporated an additional loss based on the FNN metric to penalize the encoder outputs. This formulation, however, penalizes both redundant latent variables as well as those capturing the manifold, leading to high sensitivity to the regularization term \cite{WangGuet2022}. To address this, \citet{WangGuet2022} incorporated an attention map to explicitly mask superfluous latent variables based on the FNN metric. Practically, these frameworks require repeatedly computing Euclidean distances between training data points for a range of embedding dimensions at each iteration of training, which is not ideal for systems of increasing complexity and dimensionality. Furthermore, the FNN targets the embedding dimension, which is often higher than the manifold dimension.


    Several notable methods of dimensionality reduction tools utilize local computations. A large portion of these methods belong to \KZrevise{the class of methods known as} multidimensional scaling (MDS), which are concerned with preserving some local or pairwise characteristics of the data. These include Laplacian Eigenmaps \cite{Belkin2003}, t-distributed stochastic neighbor embedding \cite{vanDerMaaten2008}, ISOMAP \cite{Tenenbaum2000} and Locally Linear Embeddings \cite{Roweis2000}. However, a major distinction between these methods and the goals of this paper is these methods require choosing a manifold dimension beforehand to embed the data into, and are generally applied towards data visualization applications. ISOMAP \cite{Tenenbaum2000}, while capable of providing an ``eyeballed'' estimate of $d_m$ via error curves, struggles to handle higher dimensionality data \cite{Camastra2016}. Several other principled $d_m$ estimation methods, such as the Levina-Bickel method \cite{LevinaBickel2004} and the Little-Jung-Maggioni method (multiscale SVD) \cite{Little2009}, estimate $d_m$ by averaging estimates made over neighborhoods of data points. {Multiscale SVD and the Levina-Bickel methods are further discussed below.} Importantly, all of these local methods lack one or more of the following features: the ability to estimate $d_m$, project new out-of-sample data points into manifold coordinates, or provide a coordinate system for the $d_m$-dimensional representation.


In this work, we address the three aforementioned challenges using a deep autoencoder framework { that drives the rank of the covariance of the data in the latent representation to a minimum. This rank will be equal to the dimension $d_m$ of the manifold where the data lies.}
Our framework utilizes two low-rank driving forces. The first is known as implicit regularization, which is a phenomenon observed in gradient-based optimization of deep linear networks (i.e. multiple linear layers in series) leading to low-rank solutions \citep{Gunasekar2017}. Although a series of linear layers is functionally and expressively identical to a single linear layer, the learning dynamics of the two are different. The mechanisms of this phenomenon are an ongoing area of research with \MDGrevise{ a primary focus on} matrix \citep{Gunasekar2017, Arora2019} and tensor factorization \citep{CohenRazin2020}. Importantly, it has been observed that implicit regularization does not occur for unstructured datasets such as random full-rank noise \citep{Arora2019}, indicating that the phenomenon depends on the underlying structure of the data. Recently, implicit regularization has been extended to autoencoders (Implicit Rank Minimizing autoencoders, IRMAE) to learn low-rank representations, improving learning representations for image-based classification, and generative problems by \citet{Jing2020}, whose observations form the foundation in this work.

The second low-rank driving force is {$L_2$ regularization, often referred to by its action when combined with gradient descent:} weight-decay. Weight-decay is a popular weight regularization mechanism in deep learning that forces the network to make trade-offs between the standard loss $L$ of the learning problem with properties of the weights of the network, $\theta$,
\begin{equation}
\mathcal{L} = L + \dfrac{\lambda}{2}  {\left\lVert \theta \right\rVert}_p^2.
\label{eq:weight_decay}
\end{equation}
Recently, \citet{mousavi-hosseini2023} showed that in two-layer neural-networks the first layer weights converge to the minimal principal subspace spanned by a target function only when online stochastic gradient descent (SGD) is combined with weight decay. The authors found that weight-decay allowed SGD to avoid critical points outside the principal subspace. Here we demonstrate a similar synergistic result when weight-decay is combined with implicitly-regularized autoencoders.


The goal of the present work is to demonstrate that implicit regularization combined with weight-decay in deep autoencoders, \KZrevise{ an approach we call \textit{Implicit Rank Minimizing Autoencoder with Weight-Decay} or \textit{IRMAE-WD}}, can be applied toward datasets that {lie on a manifold of $d_m<d_u$}, to 1) estimate the dimension of the manifold on which the data lie, 2) obtain a coordinate system describing the manifold, and 3) obtain mapping functions to and from the manifold coordinates. We highlight that IRMAE-WD produces by construction an orthogonal manifold coordinate basis organized \KZrevise{by variance},  and does not rely on extensive parameter sweeps of networks \citep{Linot2020,Vlachas2022} or external estimators \citep{WangGuet2022,Gilpin2020} -- only a good upper-bound guess of the manifold dimension is needed. (And if this guess is not good, the results of the analysis will indicate so.) These properties make the IRMAE-WD framework a natural first step for data-driven reduced-order/state space modeling and many other downstream tasks. 


\KZrevise{The remainder of this paper is organized as follows:} In Sec.~\ref{sec:Formulation} we describe the IRMAE-WD framework. In Sec.~\ref{sec:Results} we apply it to a zoo of datasets ranging from synthetic data sets to physical systems that exhibit complex chaotic dynamics including the \KZr{Lorenz system, the Kuramoto-Sivashinsky equation, and the lambda-omega reaction-diffusion system.} In Sec.~\ref{sec:robustness} we overview performance sensitivity to hyperparameters. In Sec.~\ref{sec:comparisons} we compare the framework's ability to estimate the underlying dimensionality of complex datasets against several state-of-the-art estimators. In Sec.~\ref{sec:forecasting}, we demonstrate how this framework can be naturally extended for downstream tasks such as state-space modeling and dynamics forecasting in the manifold coordinates. Finally, in Sec.~\ref{sec:dynamics}, we examine the training dynamics of IRMAE-WD to isolate the origins of low-rank in both ``space'' \KZrevise{(i.e.\ how the data representation is transformed as it passes through the architecture)} and ``time'' \KZrevise{(i.e.\ how the data representation is transformed as training progresses)}. We glean insight into network learning and, with an analysis of a special case of a linear autoencoder, provide \MDGrevise{some intuition}  for how implicit regularization and weight decay achieve a synergistic effect.  \KZrevise{Appendix \ref{sec:AppendixA} provides a summary of our architectures, Appendix \ref{sec:AppendixB} details an application to the MNIST handwriting dataset, and Appendix \ref{sec:AppendixC} contains the analysis of the linear autoencoder.} 



\section{Formulation} \label{sec:Formulation}

\begin{figure}
	\begin{center}
		\includegraphics[width=0.90\textwidth]{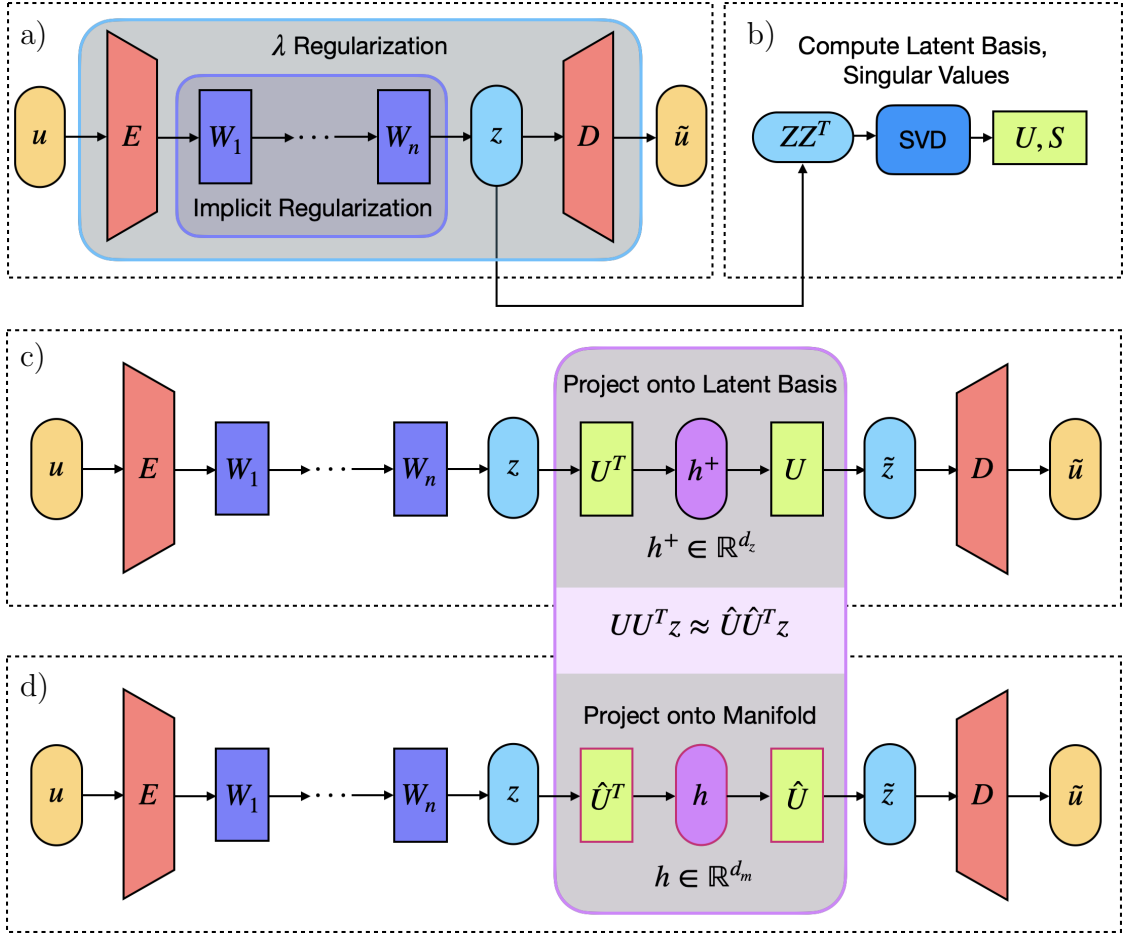}
        \captionsetup[subfigure]{labelformat=empty}
     	\begin{picture}(0,0)
    	\put(-425,337){\contour{white}{ \textcolor{black}{a)}}}
    	\put(-150,337){\contour{white}{ \textcolor{black}{b)}}}
     	\put(-425,215){\contour{white}{ \textcolor{black}{c)}}}
    	\put(-425,90){\contour{white}{ \textcolor{black}{d)}}}
    	\end{picture} 
      	\begin{subfigure}[b]{0\textwidth}\caption{}\vspace{-10mm}\label{fig:Schematic_ImplicitWD-a}\end{subfigure}
     	\begin{subfigure}[b]{0\textwidth}\caption{}\vspace{-10mm}\label{fig:Schematic_ImplicitWD-b}\end{subfigure}
        \begin{subfigure}[b]{0\textwidth}\caption{}\vspace{-10mm}\label{fig:Schematic_ImplicitWD-c}\end{subfigure}
     	\begin{subfigure}[b]{0\textwidth}\caption{}\vspace{-10mm}\label{fig:Schematic_ImplicitWD-d}\end{subfigure}
		\caption{Our implicit and $\lambda$ weight-decay regularized deep autoencoder framework a) network architecture with regularization mechanisms, b) singular value decomposition of the covariance of the learned latent data representation $Z$, c) projection of latent variables onto manifold coordinates d) isolated projection of latent variables onto manifold coordinates.}
		\label{fig:Schematic_ImplicitWD}
	\end{center}
\end{figure}

Our proposed framework uses an autoencoder architecture. Autoencoders are composed of two subnetworks, the encoder and decoder, which are connected by a latent hidden layer. For dimensionality reduction problems, this latent hidden layer is often a size-limiting bottleneck that explicitly restricts the number of degrees of freedom {available to represent}  the input data. This {architecture}, \KZrevise{which we will denote as a standard autoencoder}, forces the encoder network, $z=\mathcal{E}(u;\theta_{E})$, to compress the input data, $u\in\mathbb{R}^{d_u}$, into a compact representation, $z\in\mathbb{R}^{d_z}$, where $d_z < d_u$. The decoder, $\tilde{u}=\mathcal{D}(z;\theta_{D})$, performs the inverse task of learning to reconstruct the input, $\tilde{u}\in\mathbb{R}^{d_u}$, from the compressed representation, $z$. The autoencoder is trained to minimize the mean squared error (MSE) or reconstruction loss
\begin{equation}
\mathcal{L}(u;\theta_E,\theta_D) = \langle ||u-\mathcal{D}(\mathcal{E}(u;\theta_E);\theta_{D})||^{2}_{2} \rangle
\label{eq:AE_loss}
\end{equation}
 Here $\langle\cdot\rangle$ is the average over a training batch \KZrevise{and $\theta_i$ corresponds to the weights of each subnetwork}. We then deviate from the \KZrevise{standard autoencoder architecture} by adding an additional linear network, $\mathcal{W}(\cdot;\theta_W)$, between the encoder network and decoder: i.e.\ $z=\mathcal{W}(\mathcal{E}(u;\theta_E);\theta_{W})$, where $\mathcal{W}(\cdot;\theta_W)$ is composed of $n$ trainable linear weight matrices denoted as $W_j$ (i.e. linear layers) of size $d_z\times d_z$ in series, as was done in \citet{Jing2020}. Although $\mathcal{W}$ adds additional trainable parameters compared to a standard autoencoder, it does not give the network any additional expressivity, as linear layers in series have the same expressivity as a single linear layer. Thus, the effective capacity of the two networks are identical. \KZrevise{Importantly we train the framework with weight-decay} shown in Fig.~\ref{fig:Schematic_ImplicitWD-a} with the autoassociation task,
 \begin{equation}
\mathcal{L}(u;\theta_E,\theta_W,\theta_D) = \langle ||u-\mathcal{D}(\mathcal{W}(\mathcal{E}(u;\theta_E);\theta_{W});\theta_{D})||^{2}_{2} \rangle + \dfrac{\lambda}{2}  {\left\lVert \theta \right\rVert}_2^2.
\label{eq:our_loss}
\end{equation}
Here we contrast IRMAE-WD from typical  autoencoders tasked with finding minimal or low-dimensional representations with two distinctions. First, rather than parametrically sweep $d_z$, as is usually done with standard autoencoders, we instead guess a single $d_z > d_m$,  
and rely on implicit regularization and weight-decay to drive the latent space to an approximately minimal rank representation. {If this rank is found to equal $d_z$, then $d_z$ can be increased and the analysis repeated.}

Once the regularized network is trained, we perform singular value decomposition on the covariance matrix of the \KZrevise{latent data matrix $Z$ (i.e.\ the encoded data matrix) to obtain the matrices of singular vectors $U$ and singular values $S$}, shown in Fig.~\ref{fig:Schematic_ImplicitWD-b}. Here, the number of significant singular values of this spectrum gives an estimate of $d_m$, as each significant value represents a necessary coordinate in representing the original data in the latent space. {(More precisely, we get an estimate of $d_e$, although as we illustrate below, the analysis can be performed on subsets of data to find $d_m$ in the case $d_m<d_e$.)}

\KZrevise{Shown in Fig.~\ref{fig:Schematic_ImplicitWD-c}, we can naturally project $z$ onto $U^T$ to obtain $U^Tz = h^+ \in\mathbb{R}^{d_z}$ where each coordinate of $h^+$ is orthogonal and ordered by contribution. As $UU^T=I$, we can recover the reconstruction of $z$, $\tilde{z}$, by projecting $h^+$ onto $U$. Importantly, as the framework automatically discovers a latent space in which the encoded data only spans $d_m$ (reflected in the number of significant singular values), the data only populates the latent space in the directions of the singular vectors corresponding to those significant singular values. In other words, the encoded data does not span in the directions of the singular vectors whose corresponding singular values are approximately zero and $UU^Tz\approx \hat{U} \hat{U}^Tz$ holds, where $\hat{U}$ are the singular vectors truncated to only include those whose singular values are not approximately zero.}

\KZrevise{This observation allows us to isolate a minimal, orthogonal, coordinate system by simply projecting $z$ onto $\hat{U}$ to obtain our minimal representation $\hat{U}^Tz = h \in\mathbb{R}^{d_m}$, which we refer to as the manifold representation,
shown in Fig.~\ref{fig:Schematic_ImplicitWD-d}. As $UU^Tz\approx \hat{U} \hat{U}^Tz$ and $u \approx \mathcal{D}(\mathcal{W}(\mathcal{E}(u;\theta_E);\theta_{W});\theta_{D})$, we can transform from our manifold representation, $h$, to the ambient representation, $u$ with minimal loss.}

To glean insight into the learning mechanism of autoencoders with implicit regularization and weight-decay in a tractable manner, {in Appendix \ref{sec:AppendixC} we analyze the dynamics of gradient descent for a \emph{linear} autoencoder} acting on data whose covariance is diagonal with rank $r(=d_m)$. {For this case, there is a family of solutions for the weight matrices in which they all have rank $r$.}
 {The analysis shows that there is a ``collective" mode of decay toward the low-rank solution family in which all of the weight matrices are coupled.  The decay rate for this mode \MDGrevise{and for the mean squared error} scales as $2 + n$, where $n$ is the number of internal (square) linear layers. }In the absence of weight decay, there are directions with eigenvalues of zero that do not decay with training. When weight decay is added, these formerly zero eigenvalues become negative, allowing decay from all directions to the low-rank solution. In Sec. \ref{sec:dynamics}, we empirically observe the gradient updates and weight matrices of the linear layers of our nonlinear network exhibiting convergence \MDGrevise{toward low rank solutions}.

An important practical detail during application of the present method {is the choice of optimizer for the SGD process.} {We found that it is very important to use the AdamW optimizer \cite{loshchilov2018} rather than the standard Adam optimizer.} This distinction is important because direct application of weight decay ($L_2$ regularization) in the commonly used Adam optimizer leads to weights \KZrevise{with larger gradient amplitudes} being regularized \KZrevise{disproportionately} \cite{loshchilov2018}. AdamW decouples weight decay from the adaptive gradient update. We have found that the usage of the base Adam optimizer with $L_2$ regularization can lead to high sensitivity to parameters and spurious results.

\section{Results} 
\subsection{Manifold Dimension Estimates: Example Systems} \label{sec:Results}
We now investigate IRMAE-WD applied to a zoo of datasets of increasing complexity, ranging from linear manifolds embedded in finite-dimensional ambient spaces to nonlinear manifolds embedded in formally infinite-dimensional ambient spaces.

\subsubsection{Data linearly embedded in a finite-dimensional ambient space}
\begin{figure}
	\begin{center}
		\includegraphics[width=0.90\textwidth]{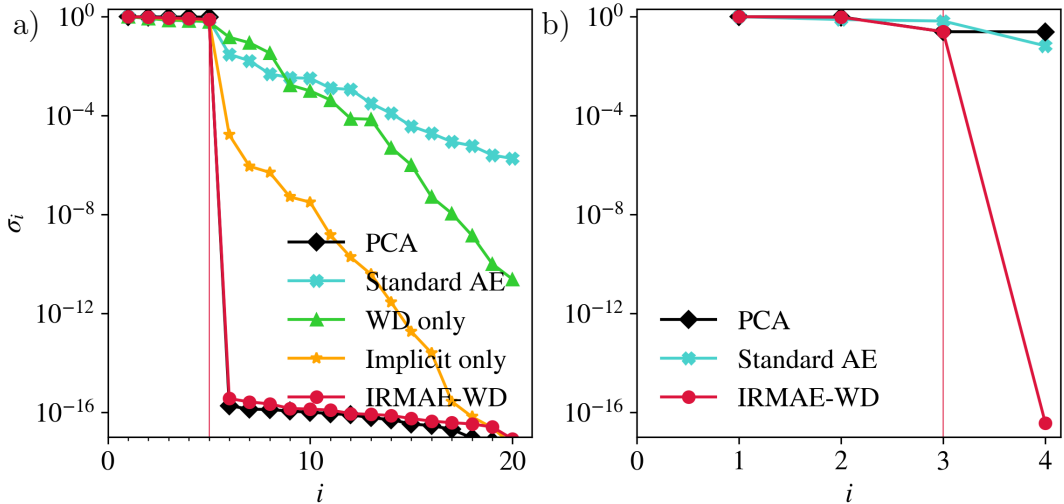}
        \captionsetup[subfigure]{labelformat=empty}
        \begin{picture}(0,0)
    	\put(-415,190){\contour{white}{ \textcolor{black}{a)}}}
    	\put(-215,190){\contour{white}{ \textcolor{black}{b)}}}
    	\end{picture} 
      	\begin{subfigure}[b]{0\textwidth}\caption{}\vspace{-10mm}\label{fig:SV_Ortho_Arch-a}\end{subfigure}
     	\begin{subfigure}[b]{0\textwidth}\caption{}\vspace{-10mm}\label{fig:SV_Ortho_Arch-b}\end{subfigure}
		\caption{Normalized singular values, $\sigma_i$, of {latent data covariances} of various AE methods applied to a) a 5-dimensional linear manifold embedded in $\mathbb{R}^{20}$ and b) a 3-dimensional nonlinear manifold embedded in $\mathbb{R}^{4}$. The spectra obtained from PCA and a standard AE with no regularization are provided. The {value of } $d_m$ is marked by the vertical red guide line.}
		\label{fig:SV_Ortho_Arch}
	\end{center}
\end{figure}

We first benchmark IRMAE-WD against a simple data set consisting of 5-dimensional noise linearly embedded in an ambient space of 20 dimensions. Because this dataset exactly spans 5 orthogonal directions and is linearly embedded, Principal Component Analysis (PCA) is able to extract $d_m$ from the data, which can be identified via the singular value spectrum of \KZrevise{covariance of the data matrix}. Shown in Fig.~\ref{fig:SV_Ortho_Arch-a} are the singular values obtained from PCA, from the learned latent variables of IRMAE without and with weight-decay, and a standard AE \KZrevise{that is architecturally identical to IRMAE-WD without any regularization (i.e.\ no $W$ and $\lambda=0$)}.
\KZrevise{For the standard autoencoder, while the singular values $\sigma_i$  drop slightly for index $i>5$, the spectrum is broad and decays slowly, indicating that the learned latent representation is essentially full-rank. In other words, the standard autoencoder, when given excess capacity in the bottleneck layer, will utilize all latent variables available to it. In contrast, for IRMAE-WD, the singular values for $i>5$ drop to $\sim 10^{-16}$, just as in the case of PCA. This indicates that IRMAE-WD is able to automatically learn a representation that isolates the minimal dimensions needed to represent the data.}

We further highlight here two important observations: 1) an autoencoder with weight decay alone is insufficient in learning a sparse representation -- it behaves very similarly to the standard autoencoder, and 2) an autoencoder with implicit regularization alone, as applied in \citet{Jing2020}, yields a sharp drop in $\sigma_i$ for $i>5$, but not nearly so dramatic as when both linear layers and weight decay are implemented. {This phenomenon is addressed in Sections \ref{sec:dynamics} and Appendix \ref{sec:AppendixC}.}


\subsubsection{Nonlinearly embedded finite-dimensional system: The Archimedean Spiral Lorenz}

We now turn our attention to data from nonlinear finite-dimensional dynamical systems with nonlinear embedded manifolds. Specifically, we take the Lorenz `63 system \cite{Lorenz1963},
\begin{equation}
\begin{split}
\dot{x} &= \sigma(y-x) \\
\dot{y} &= x(\rho-z)-y \\
\dot{z} &= xy-\beta z
\end{split}
\label{eq:Lorenz}
\end{equation}
which exhibits chaotic dynamics in $\mathbb{R}^3$ and embed this system nonlinearly in $\mathbb{R}^4$ by wrapping the data set around the Archimedean spiral using the following mapping: $$[x, y, \alpha z \cos{(\alpha z)}, \alpha z \sin{(\alpha z)}]\rightarrow [u_1, u_2, u_3, u_4],$$  with $\alpha=0.2$. 

\begin{figure}
	\begin{center}
		\includegraphics[width=0.90\textwidth]{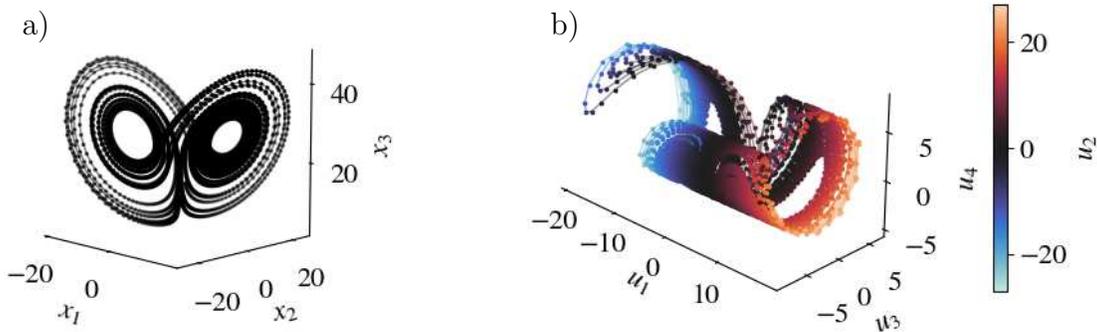}
        \begin{picture}(0,0)
    	\put(-420,120){\contour{white}{ \textcolor{black}{a)}}}
    	\put(-220,120){\contour{white}{ \textcolor{black}{b)}}}
    	\end{picture} 
		\caption{Dynamics of the a) 3-dimensional Lorenz `63 equation and b) the 4-dimensional Archimedean Lorenz equation. \KZrevise{The color corresponds to the variable, $u_2$ in the embedding, while the spatial coordinates correspond to $u_1$, $u_3$, and $u_4$.}}
		\label{fig:Data_ArchLorenz}
	\end{center}
\end{figure}

For parameters $\sigma=10,\rho=28,\beta=8/3$,  the Lorenz `63 exhibits chaotic dynamics. In other words, the underlying dynamics of this system live on a nonlinear 3-dimensional manifold that is nonlinearly embedded in a 4-dimensional ambient space. We show in Fig.~\ref{fig:SV_Ortho_Arch-b} that IRMAE-WD correctly determines that this system can be minimally represented by 3 latent variables.  In contrast, the application of PCA fails to identify the underlying structure of the data. Here, the PCA spectrum does not give a correct estimate of $d_m$ because inherently a linear method cannot minimally capture the nonlinearity/curvature of the manifold. Finally, a standard AE with $d_z > d_m$ also fails to automatically learn a minimal representation as it finds a full-rank {data covariance in the }latent space.


\subsubsection{Global manifold estimates vs local estimates: quasiperiodic dynamics on a 2-torus}
\begin{figure}
	\begin{center}
		\includegraphics[width=0.90\textwidth]{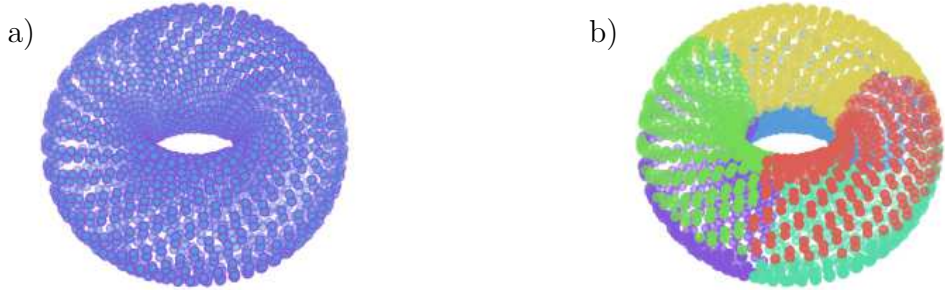}
        \captionsetup[subfigure]{labelformat=empty}
        \begin{picture}(0,0)
    	\put(-400,140){\contour{white}{ \textcolor{black}{a)}}}
    	\put(-180,140){\contour{white}{ \textcolor{black}{b)}}}
    	\end{picture} 
    	\begin{subfigure}[b]{0\textwidth}\caption{}\vspace{-10mm}\label{fig:Data_Torus2D-a}\end{subfigure}
     	\begin{subfigure}[b]{0\textwidth}\caption{}\vspace{-10mm}\label{fig:Data_Torus2D-b}\end{subfigure}
		\caption{Quasiperiodic dynamics on a torus: a) global b) local patches.}
		\label{fig:Data_Torus2D}
	\end{center}
\end{figure}

\begin{figure}
	\begin{center}
		\includegraphics[width=0.90\textwidth]{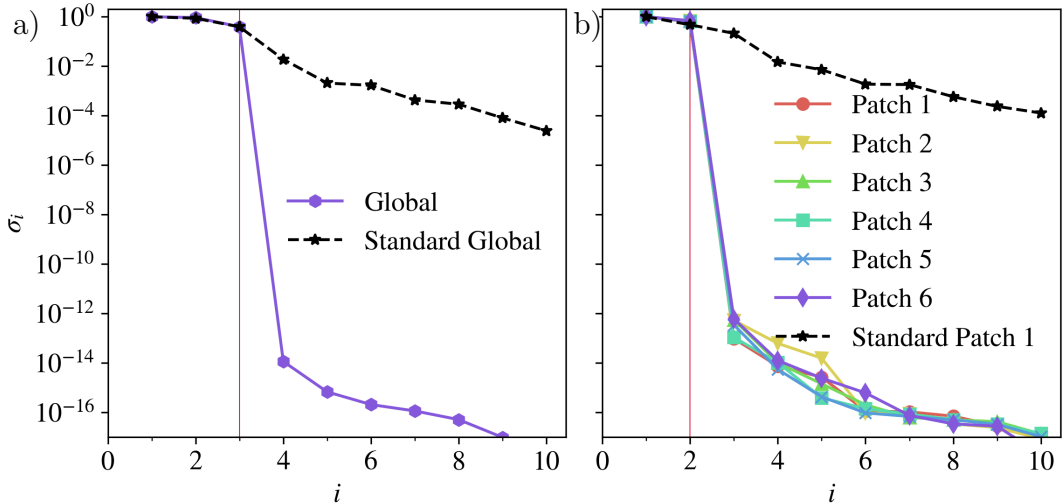}
        \captionsetup[subfigure]{labelformat=empty}
        \begin{picture}(0,0)
    	\put(-415,190){\contour{white}{ \textcolor{black}{a)}}}
    	\put(-203,190){\contour{white}{ \textcolor{black}{b)}}}
    	\end{picture} 
    	\begin{subfigure}[b]{0\textwidth}\caption{}\vspace{-10mm}\label{fig:SV_Torus2D-a}\end{subfigure}
     	\begin{subfigure}[b]{0\textwidth}\caption{}\vspace{-10mm}\label{fig:SV_Torus2D-b}\end{subfigure}
		\caption{Normalized singular values, $\sigma_i$, of learned latent spaces from IRMAE-WD applied to the a) global torus dataset and b) local patches of the torus dataset. Results for a standard AE latent space is shown in black. The value of $d_m$ is marked by the vertical red guide line.}
		\label{fig:SV_Torus2D}
	\end{center}
\end{figure}

We now turn our attention to a trajectory in $\mathbb{R}^3$ traversing the surface of a 2-torus with poloidal and toroidal speeds that lead to quasiperiodic dynamics, as visualized in Fig.~\ref{fig:Data_Torus2D-a}. Given infinite time, the particle will densely cover the surface of the torus. Although this system lives on a two-dimensional manifold, {the topology of this manifold is nontrivial and} a single global representation is not possible to obtain \cite{FloryanCANDyMan2022}. Here we apply IRMAE-WD to this dataset, which consists of snapshots of the three coordinates along a trajectory. In this example, we also use an overcomplete network design, $z\in\mathbb{R}^{10}$, to highlight that even when $d_z>d_u$, excess degrees of freedom are still correctly {eliminated}.
We show in Fig.~\ref{fig:SV_Torus2D-a} that when IRMAE-WD is tasked with learning a global representation by training over the entire dataset, it (correctly) obtains a 3-dimensional latent space -- the embedding dimension of the manifold is $d_e=3$. However, as described by \citet{FloryanCANDyMan2022}, by decomposition of the manifold into an atlas of overlapping charts, the intrinsic dimension of the manifold containing the data can be captured. In Fig.~\ref{fig:SV_Torus2D-b}, we show IRMAE-WD applied to the same dataset after being divided into \KZrevise{patches found using k-means clustering}, illustrated in Fig.~\ref{fig:Data_Torus2D-b}. We show that for each subdomain, IRMAE-WD automatically learns a minimal 2-dimensional representation of the data while simultaneously discarding the remaining superfluous degrees of freedom. 
\KZrevise{In this manner, IRMAE-WD can be deployed on local regions of data to make estimates of the intrinsic dimension.} 

\subsubsection{Nonlinear manifold in an ``infinite-dimensional"  system: Kuramoto-Sivashinsky equation and reaction-diffusion system}

\begin{figure}
	\begin{center}
		\includegraphics[width=0.95\textwidth]{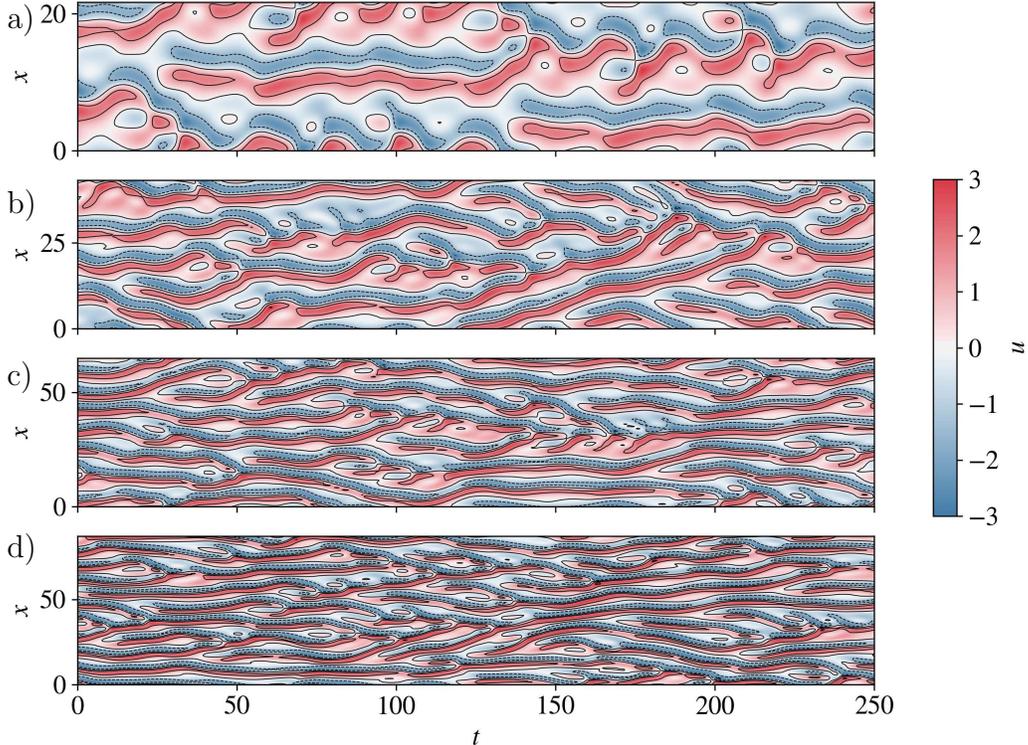}
        \captionsetup[subfigure]{labelformat=empty}
     	\begin{picture}(0,0)
    	\put(-425,285){\contour{white}{ \textcolor{black}{a)}}}
    	\put(-425,215){\contour{white}{ \textcolor{black}{b)}}}
     	\put(-425,150){\contour{white}{ \textcolor{black}{c)}}}
    	\put(-425,85){\contour{white}{ \textcolor{black}{d)}}}
    	\end{picture} 
      	\begin{subfigure}[b]{0\textwidth}\caption{}\vspace{-10mm}\label{fig:Data_KSE_22_44_66_88-a}\end{subfigure}
     	\begin{subfigure}[b]{0\textwidth}\caption{}\vspace{-10mm}\label{fig:Data_KSE_22_44_66_88-b}\end{subfigure}
        \begin{subfigure}[b]{0\textwidth}\caption{}\vspace{-10mm}\label{fig:Data_KSE_22_44_66_88-c}\end{subfigure}
     	\begin{subfigure}[b]{0\textwidth}\caption{}\vspace{-10mm}\label{fig:Data_KSE_22_44_66_88-d}\end{subfigure}
		\caption{Typical evolutions for the KSE in domain sizes of a) $L=22$, b) $L=44$, c) $L=66$, and d) $L=88$.}
		\label{fig:Data_KSE}
	\end{center}
\end{figure}

\begin{figure}
	\begin{center}
		\includegraphics[width=0.95\textwidth]{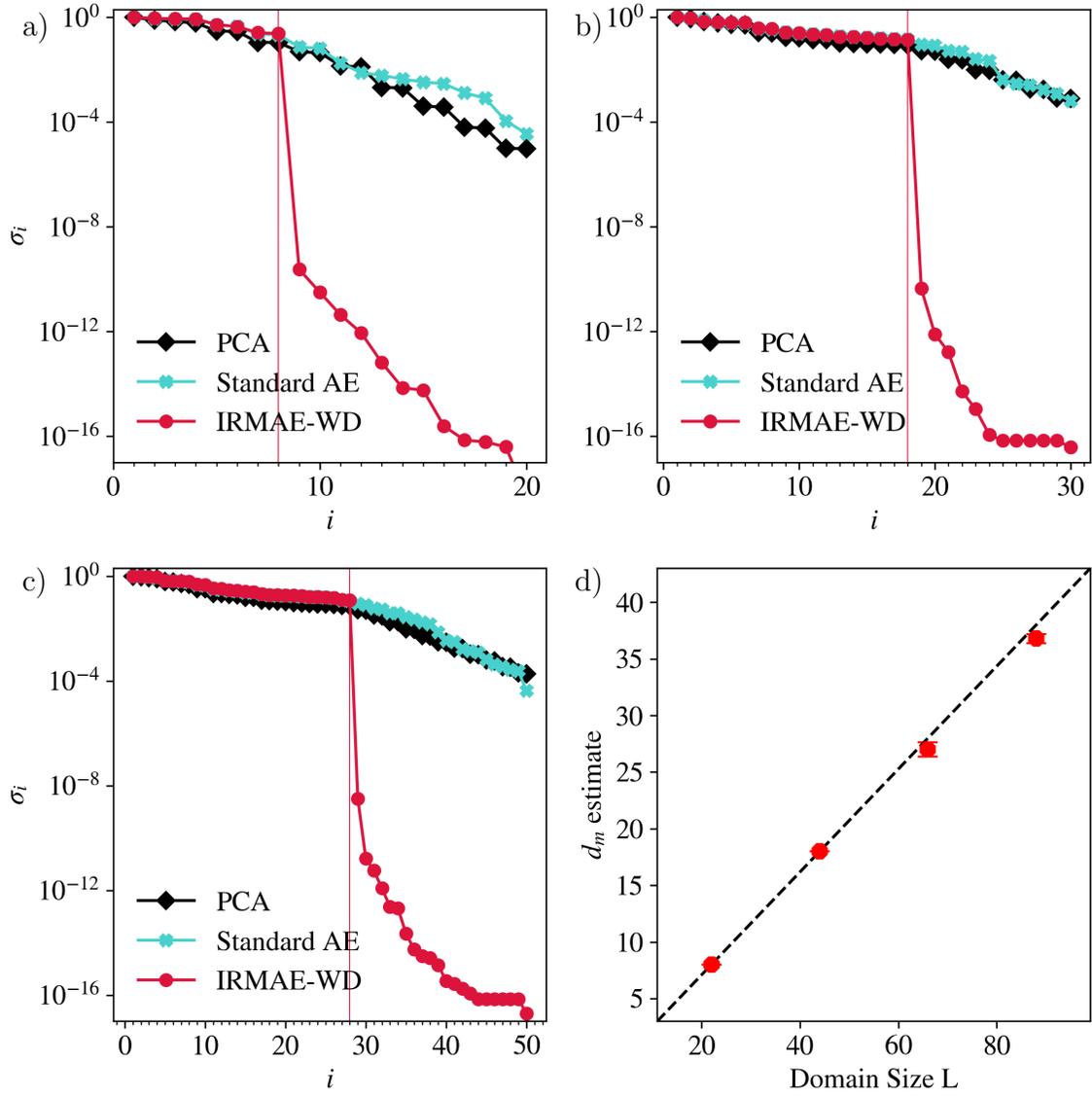}
        \captionsetup[subfigure]{labelformat=empty}
        \begin{picture}(0,0)
    	\put(-435,423){\contour{white}{ \textcolor{black}{a)}}}
    	\put(-220,423){\contour{white}{ \textcolor{black}{b)}}}
     	\put(-435,207){\contour{white}{ \textcolor{black}{c)}}}
    	\put(-220,207){\contour{white}{ \textcolor{black}{d)}}}
    	\end{picture} 
    	\begin{subfigure}[b]{0\textwidth}\caption{}\vspace{-10mm}\label{fig:SV_KSE_L22_L44_L66_L88-a}\end{subfigure}
     	\begin{subfigure}[b]{0\textwidth}\caption{}\vspace{-10mm}\label{fig:SV_KSE_L22_L44_L66_L88-b}\end{subfigure}
        \begin{subfigure}[b]{0\textwidth}\caption{}\vspace{-10mm}\label{fig:SV_KSE_L22_L44_L66_L88-c}\end{subfigure}
     	\begin{subfigure}[b]{0\textwidth}\caption{}\vspace{-10mm}\label{fig:SV_KSE_L22_L44_L66_L88-d}\end{subfigure}
		\caption{Singular values, $\sigma_i$, of IRMAE-WD learned latent spaces for the KSE a) $L=22$, b) $L=44$, c) $L=66$, and d) \KZrevise{estimate of $d_m$  averaged over 5 randomly initialized models as a function of $L$, with the standard deviations represented by the error bars}. In a)-c), the spectra obtained from PCA and a standard AE with no regularization are also shown, and the value of $d_m$ is marked by the vertical red guide line.}
		\label{fig:SV_KSE_L22_L44_L66_L88}
	\end{center}
\end{figure}

We now turn our attention to two dissipative nonlinear systems that are formally infinite-dimensional -- nonlinear PDEs -- which are discretized to become high-dimensional systems of ODEs. First, we investigate the 1D Kuramoto-Sivashinsky equation (KSE):
\begin{equation}
\frac{\partial v}{\partial t} = -v\frac{\partial v}{\partial x} -\frac{\partial^2 v}{\partial x^2} -\frac{\partial^4 v}{\partial x^4}
\label{eq:KSE}
\end{equation}
{in a domain of length $L$ with periodic boundary conditions.}
For large $L$, this system exhibits rich spatiotemporal chaotic dynamics which has made it a common test case for studies of complex nonlinear systems. To analyze this formally ``infinite'' dimensional system, state snapshots will consist of sampled solution values at equidistant mesh points in the domain. We apply IRMAE-WD to extract the dimension of the underlying manifold for dynamics for { a range of domain sizes, focusing first on} $L=22$, which exhibits spatiotemporal chaotic dynamics {and has been widely studied}. An example trajectory of this system is shown in Fig.~\ref{fig:Data_KSE_22_44_66_88-a}. This system, although formally infinite-dimensional, has dynamics dictated by a nonlinearly embedded 8-dimensional manifold, as indicated by a variety of methodologies \citep{Ding2016,Takeuchi2011,Yang2009,Linot2020}. Using a data set comprised of 40,000 snapshots sampled on 64 mesh points, and choosing a bottleneck layer dimension $d_z=20$, we show in Fig.~\ref{fig:SV_KSE_L22_L44_L66_L88-a} that the singular values coming from IRMAE-WD drop dramatically above an index of 8, indicating that we have automatically and straightforwardly learned a latent space of dimension $d_m=8$. {By contrast, neither PCA nor a standard AE leads to a substantial drop in singular values over the whole range of indices. }

For increasing domain sizes \MDGrevise{$L$}, the spatiotemporal dynamics of the \MDGrevise{KSE} increase in complexity. Fig.~\ref{fig:Data_KSE}b-d show space-time plots of the dynamics for $L=44,66$, and $88$, sampled on a uniform spatial mesh of 64, 64, and 128 points, respectively. Fig.~\ref{fig:SV_KSE_L22_L44_L66_L88}b-c show the singular value spectra of the latent space covariances for $L=44$ and $66$, again showing a drop of $>10$ orders of magnitude at well-defined index values, indicating manifold dimensions $d_m=18$ and $28$, respectively.   We highlight here that previous autoencoder methods \citep{Linot2020,Linot2021}, using the trend in MSE with $d_z$ to estimate $d_m$, struggle to make distinctions in the manifold dimension for these domain sizes, while IRMAE-WD yields a well-characterized value. Prior works relying on high-precision analyses of the dynamics based on detailed and complex trajectory analyses have suggested that the manifold dimension for the KSE scales linearly with the domain length $L$ \citep{Yang2009,Takeuchi2011}. In Fig.~\ref{fig:SV_KSE_L22_L44_L66_L88-d} we show the trend in $d_m$ vs.~$L$ as determined with IRMAE-WD: we are able to very straightforwardly recover the linear scaling without access to the underlying governing equations or periodic solutions. 





\begin{figure}
	\begin{center}
		\includegraphics[width=0.5\textwidth]{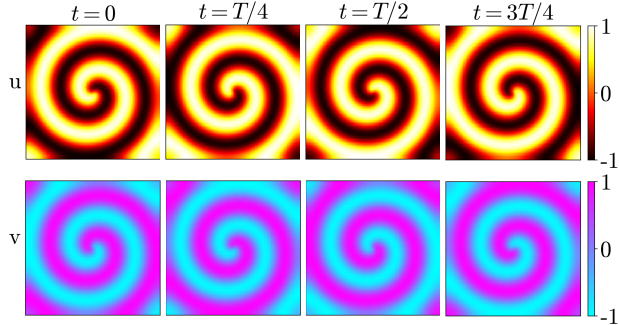}
        \captionsetup[subfigure]{labelformat=empty}
        \begin{picture}(0,0)
    	\end{picture} 
		\caption{One period $T$ of the spiral wave produced by lambda-omega reaction diffusion system.}
		\label{fig:ReactionDiffusionExample}
	\end{center}
\end{figure}

\begin{figure}
	\begin{center}
		\includegraphics[width=0.5\textwidth]{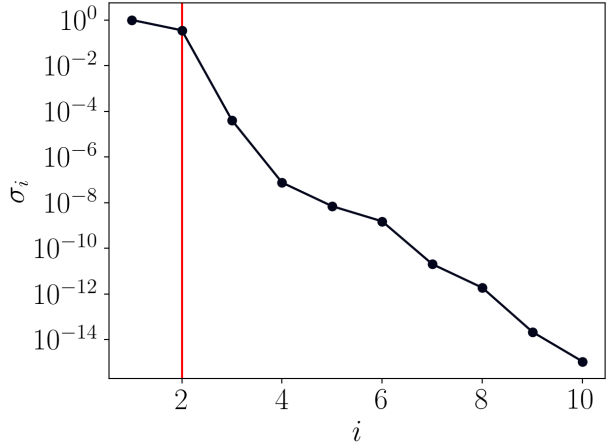}
        \captionsetup[subfigure]{labelformat=empty}
        \begin{picture}(0,0)
    	\end{picture} 
		\caption{Singular values $\sigma_i$ of IRMAE-WD learned latent space covariance for the lambda-omega reaction-diffusion system. }
		\label{fig:ReactionDiffusionIRMAE}
	\end{center}
\end{figure}

The second infinite-dimensional system we consider is the lambda-omega reaction-diffusion system in two spatial dimensions governed by

\begin{equation}
    \begin{split}
        \frac{\partial u}{\partial t} & = \left[ 1 - \left( u^2 +v^2 \right) \right] u + \beta \left( u^2 + v^2 \right) v + d_1 \left( \frac{\partial^2 u}{\partial x^2} +\frac{\partial^2 u}{\partial y^2} \right)\\
        \frac{\partial v}{\partial t} & = - \beta \left( u^2 + v^2 \right) u + \left[ 1 - \left( u^2 +v^2 \right) \right] v + d_2 \left( \frac{\partial^2 u}{\partial x^2} +\frac{\partial^2 u}{\partial y^2} \right)
    \end{split}
\label{eq:Reaction-diffusion}
\end{equation}

where $d_1 = d_2 = 0.1$ and $\beta = 1$ for $-10 \leq x \leq 10, -10 \leq y \leq 10$. This system has previously been studied in \citep{Champion2019,FloryanCANDyMan2022,Kicic.2023.10.1016/j.cma.2023.116204}. The long-time dynamics of the system collapse onto an attracting limit cycle in state space in the form of a spiral wave, which can be fully described in a 2-dimensional latent space with a single global representation \citep{FloryanCANDyMan2022}. We will analyze this system with state snapshots sampled from solution values at equidistant mesh points in a $101 \times 101$ grid, producing an ambient dimension of $\mathbb{R}^{20402}$. We generated a data set comprised of 201 snapshots, uniformly spaced 0.05 time units apart, covering slightly over one period of the spiral wave; one period of the spiral wave is shown in Fig.~\ref{fig:ReactionDiffusionExample}. We applied IRMAE-WD using a bottleneck layer dimension of $d_z = 10$, and we show in Fig.~\ref{fig:ReactionDiffusionIRMAE} that the singular values drastically decrease above an index of 2, indicating that we have automatically and straighforwardly learned a latent space of dimension $d_m = 2$.

\subsection{Robustness and Parameter Sensitivity} \label{sec:robustness}

In the following section we overview parametric robustness of IRMAE-WD, focusing on the KSE $L=22$ dataset. We choose this dataset as it comes from a nonlinear, high-dimensional system governed by dynamics on a nonlinear manifold and is considerably more complex than typical benchmark systems. 

\begin{figure}
	\begin{center}
		\includegraphics[width=0.95\textwidth]{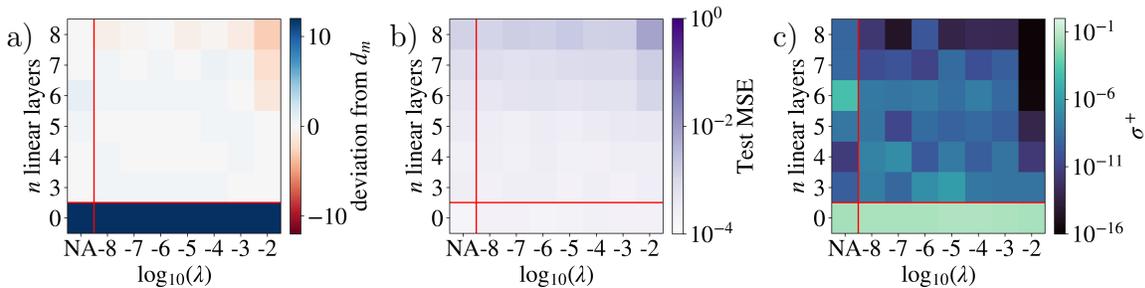}
        \captionsetup[subfigure]{labelformat=empty}
        \begin{picture}(0,0)
    	\put(-450,110){\contour{white}{ \textcolor{black}{a)}}}
    	\put(-305,110){\contour{white}{ \textcolor{black}{b)}}}
        \put(-160,110){\contour{white}{ \textcolor{black}{c)}}}
    	\end{picture} 
        \begin{subfigure}[b]{0\textwidth}\caption{}\vspace{-10mm}\label{fig:KSE_Parametric_MSE_dm_svf-a}\end{subfigure}
     	\begin{subfigure}[b]{0\textwidth}\caption{}\vspace{-10mm}\label{fig:KSE_Parametric_MSE_dm_svf-b}\end{subfigure}
        \begin{subfigure}[b]{0\textwidth}\caption{}\vspace{-10mm}\label{fig:KSE_Parametric_MSE_dm_svf-c}\end{subfigure}
		\caption{Parametric sweep in $n$ and $\lambda$ of models trained over the KSE $L=22$ dataset with various degrees of implicit and weight regularization. The colors correspond to: a) average deviation from $d_m$ over 3 models, b)  average test MSE over 3 models, c) lowest  fraction of trailing singular values of 3 models (lower is better). In the leftmost column, labeled NA, $\lambda=0$. The lower left corner in each plot corresponds to a standard autoencoder. }
		\label{fig:Parametric_Sweep}
	\end{center}
\end{figure}

We first investigate the accuracy of the estimate of $d_m${, where the correct value, based on consistent results from many sources, is taken to be $d_m=8$.} {Fig.~\ref{fig:KSE_Parametric_MSE_dm_svf-a} shows the dimension estimate as a function of number of linear layers $n$ and weight decay parameter $\lambda$, with the bottom row of the plot corresponding to the case $n=0$ of a standard autoencoder with $L_2$ regularization.} We highlight that for a broad range of $n$ and $\lambda$ the framework is capable of accurately estimating $d_m$. It is not until there is significant regularization {in terms of both $n$ and $\lambda$} that the framework begins to fail. In the absence of implicit regularization with linear layers the autoencoder cannot predict $d_m$ at all. Shown in Fig.~\ref{fig:KSE_Parametric_MSE_dm_svf-b} is the same parameter sweep characterized by test MSE performance. This quantity is also relatively insensitive to choice of parameters,  and the regularized models operating with effectively $d_m$ degrees of freedom in the representation achieve comparable reconstruction errors to standard autoencoders (bottom left corner). Finally, for an ideal regularized model, singular values with indices greater than $d_m$ are zero, but practically this is not the case. In Fig.~\ref{fig:KSE_Parametric_MSE_dm_svf-c}, we quantify the fraction of total variance in the representation coming from singular values from the tail of the spectrum, i.e.~with index greater than $d_m$: $\sigma^+=\sum_{i=d_{m+1}}^{d_z}\sigma_i / \sum_{j=1}^{d_z}\sigma_j $.
We highlight here that for a broad range of $n$ and $\lambda$, the trailing singular values contribute on the order of $10^{-9}$ of the total variance or energy, while the unregularized models contribute a nontrivial $10^{-1}$. Finally, we comment that we did not observe strong dependence {of the choice of $d_z$ on the results, as long as $d_z>d_m$.}

\subsection{Comparison to other methods}
\label{sec:comparisons}
\begin{figure}
	\begin{center}
		\includegraphics[width=0.8\textwidth]{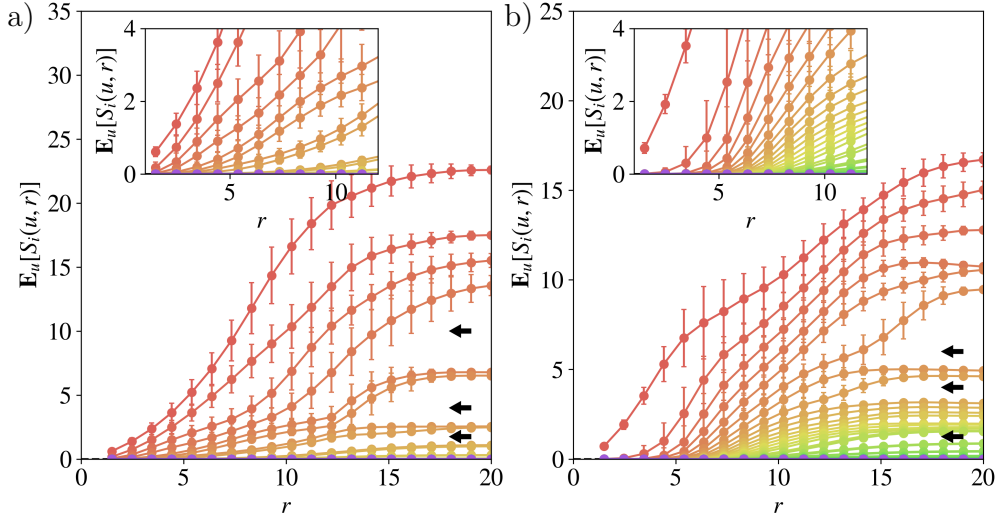}
        \captionsetup[subfigure]{labelformat=empty}
        \begin{picture}(0,0)
    	\put(-385,190){\contour{white}{ \textcolor{black}{a)}}}
    	\put(-197,190){\contour{white}{ \textcolor{black}{b)}}}
    	\end{picture} 
    	\begin{subfigure}[b]{0\textwidth}\caption{}\vspace{-10mm}\label{fig:MSVD_KSE_L22_L44-a}\end{subfigure}
     	\begin{subfigure}[b]{0\textwidth}\caption{}\vspace{-10mm}\label{fig:MSVD_KSE_L22_L44-b}\end{subfigure}
		\caption{Ensemble MSVD singular values, $S_i$, as a function of sampling neighborhood radius $r$ on the KSE a) $L=22$ dataset and b) $L=44$ dataset. The color of the lines corresponds to the modal index of the spectra. The arrows mark the gaps that appear in the spectra, providing an estimate of underlying dimensionality.}
		\label{fig:MSVD_KSE_L22_L44}
	\end{center}
\end{figure}

In this section, we compare IRMAE-WD to two state-of-the-art estimators: Multiscale SVD (MSVD) \cite{Little2009} and the Levina-Bickel method \cite{LevinaBickel2004} \KZrevise{as these methods are designed to provide a direct estimate of the manifold dimension from data}. 
\KZrevise{We first compare to the MSVD method, as it is also completely data-driven and also relies on analyzing singular value spectra of the data. MSVD estimates $d_m$ by tracking the ensemble average of singular value spectra obtained from a collection of local neighborhoods of data as a function of the size of the neighborhood radius, $r$. Development of gaps in the singular value spectra  as $r$ increases coincides with a separation between directions on the manifold and those due to curvature. {A gap in the spectrum is presumed to indicate the manifold dimension, as PCA does for data on a linear manifold.} We revisit the KSE $L=22$ and $L=44$ datasets as they are nontrivial complex systems with nonlinear manifolds.} In Fig.~\ref{fig:MSVD_KSE_L22_L44-a} and Fig.~\ref{fig:MSVD_KSE_L22_L44-b} we show the MSVD method applied to these datasets. 
We highlight here that MSVD, given these datasets, is unable to unambiguously identify $d_m$; {Rather than one gap, there are multiple gaps in the spectra, as indicated by the arrows.} This is likely due to a key limitations of MSVD, which is that it requires data in small enough $r$ neighborhoods to accurately approximate the highly nonlinear manifold as flat.
I.e.~in order to work in the limit of very small neighborhoods, MSVD requires an ensemble of data points to have {a sufficient number of} neighboring points at very small $r$. In our MSVD application, we were unable to access small values of $r$ without encountering neighborless point cloud samples. Many complex dynamical systems do not uniformly populate their underlying manifolds, resulting in regions of high and low density  -- indeed,  data points on a chaotic attractor will be fractally, rather than uniformly, distributed. As a result, it is difficult to collect dynamical data in which the manifold is represented with uniform density or to collect enough data such that low probability regions are dense when natural occurrences in these regions are low. {IRMAE-WD does not suffer from these limitations.}

\KZrevise{We also apply the Levina-Bickel method to these same datasets. This method utilizes a maximum likelihood framework in estimating the dimensionality of the data from local regions \citep{LevinaBickel2004}. In our application we fix the number of neighbors, as suggested by \citet{LevinaBickel2004}, rather than fixing the neighborhood radius.} This method also fails to provide reliable estimates given our datasets. We summarize this section with our findings in Table~\ref{table:dm_methods}. For the datasets considered, the Levina-Bickel method appears to underestimate the dimensionality while MSVD tends to give ambiguous estimates.

\begin{table}[ht]
\caption{Estimates of $d_m$ with various methods.} 
\centering 
\begin{tabular}{c c c c c} 
\hline\hline 
Dataset & $d_m$ & Multiscale SVD & Levina-Bickel & IRMAE-WD \\ [0.5ex] 
\hline 
Arch. Lorenz & 3 & 2 & 2.09 & 3 \\
KSE $L=22$ & 8 & 6-8 & 3.99 & 8 \\ 
KSE $L=44$ & 18 & 8-20 & 7.00 & 18 \\ [1ex] 
\hline 
\end{tabular}
\label{table:dm_methods} 
\end{table}

\subsection{Reduced-order state-space forecasting in the manifold coordinates} \label{sec:forecasting}



\begin{figure}[t]
	\begin{center}
		\includegraphics[width=0.95\textwidth]{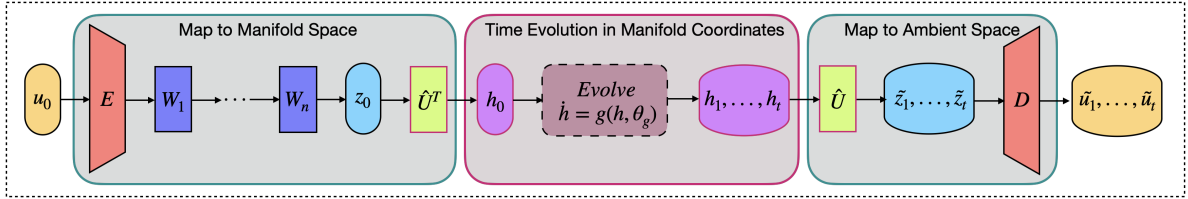}
		\caption{Schematic for extending the IRMAE-WD framework for forecasting in the manifold coordinate system using a Neural ODE (pink section). }
		\label{fig:Schematic_Forecasting}
	\end{center}
\end{figure}

{As noted above and illustrated in Fig.~\ref{fig:Schematic_ImplicitWD}, projection of the latent space data $z$ onto the first $d_m$ singular vectors of its covariance yields the manifold representation $h\in\mathbb{R}^{d_m}$. }
We can map data snapshots in the ambient space to this manifold coordinate representation by simply extending our definitions of encoding and decoding to $h$:
\begin{equation}
    \begin{split}
        h & := \mathcal{E}_h(u;\theta_E,\theta_W,\hat{U}^T) = \hat{U}^T \mathcal{W}(\mathcal{E}(u;\theta_E);\theta_W) \\
        \tilde{u} & := \mathcal{D}_h(h;\theta_D,\hat{U}^T) = \mathcal{D}(\hat{U}h;\theta_D)
    \end{split}
\label{eq:encode_h}
\end{equation}
where $\mathcal{E}_h$ and $\mathcal{D}_h$ simply subsume the intermediate linear transformations required to map between $h$ and $u$. With our extended definitions of encoding and decoding, we now have 1) found an estimate of $d_m$, 2) obtained the coordinate system $h$ parameterizing the manifold, and 3) determined the explicit mapping functions $\mathcal{E}_h$ and $\mathcal{D}_h$ back and forth between the ambient space and data manifold. With access to these three, a natural application is state-space modeling and forecasting. We show a schematic of this extension in Fig.~\ref{fig:Schematic_Forecasting}; the pink internal box contains a time-evolution module to integrate an initial condition $u_0$ that has been transformed into manifold coordinate representation $h_0$ forward in time.  \MDGrevise{Having in hand an explicit determination of the manifold dimension and coordinates, it is now no longer necessary to use trial and error, testing models with various dimensions (as in~e.g.~\cite{Linot2021}), to find a minimal-dimensional high-fidelity time-evolution model.}

\MDGrevise{Before continuing to examples, we make some general comments about the approach and setting addressed in this section. We are considering deterministic dynamical systems with long-time dynamics that lie on an invariant manifold of dimension smaller than the ambient dimension. A simple example would be a system with a stable limit cycle. Topologically this is one-dimensional but its embedding dimension is two. That means that two global coordinates, \emph{no more and no fewer}, are necessary and sufficient for prediction of the dynamics on the limit cycle.  Our aim is to identify these coordinates and the dynamics in them.  In a system whose exact (manifold) representation requires a large number of coordinates, it may still be possible to develop a model that predicts many aspects of the system, especially with regard to statistics, with a model that has many fewer dimensions than the true invariant manifold. That is a common goal, and is for example what is done in large eddy simulations of turbulent flow \cite{Bose.2018.10.1146/annurev-fluid-122316-045241}. But that is not what we are aiming to do here. }

\begin{figure}
	\begin{center}
		\includegraphics[width=0.95\textwidth]{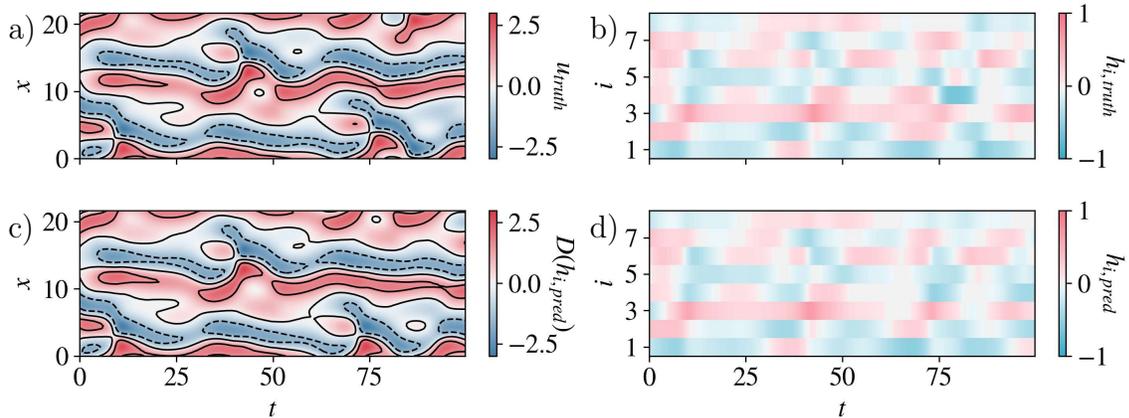}
        \captionsetup[subfigure]{labelformat=empty}
        \begin{picture}(0,0)
    	\put(-445,155){\contour{white}{ \textcolor{black}{a)}}}
    	\put(-225,155){\contour{white}{ \textcolor{black}{b)}}}
        \put(-445,80){\contour{white}{ \textcolor{black}{c)}}}
    	\put(-225,80){\contour{white}{ \textcolor{black}{d)}}}
    	\end{picture} 
    	\begin{subfigure}[b]{0\textwidth}\caption{}\vspace{-10mm}\label{fig:Latent_KSE_Forecasting-a}\end{subfigure}
     	\begin{subfigure}[b]{0\textwidth}\caption{}\vspace{-10mm}\label{fig:Latent_KSE_Forecasting-b}\end{subfigure}
        \begin{subfigure}[b]{0\textwidth}\caption{}\vspace{-10mm}\label{fig:Latent_KSE_Forecasting-c}\end{subfigure}
     	\begin{subfigure}[b]{0\textwidth}\caption{}\vspace{-10mm}\label{fig:Latent_KSE_Forecasting-d}\end{subfigure}
		\caption{Example ground truth trajectory of the KSE in the a) ambient space and  b) projected onto the learned manifold coordinate representation. A time series prediction made using a Neural ODE in the d) manifold coordinate beginning from the same initial condition. c) The ambient space reconstruction decoded from the neural ODE predicted manifold trajectory.}
		\label{fig:Latent_KSE_Forecasting}
	\end{center}
\end{figure}

\MDGrevise{We now extend the KSE and reaction-diffusion examples described above to develop data-driven dynamical models in the manifold coordinates.}
Here we train a neural ODE \cite{Chen2018}, $\dot{h}=g(h;\theta_g)$, to model the time evolution of $h$ \KZrevise{as done by \citet{Linot2021}}. In other words, we simply insert a forecasting network trained to evolve the dynamics of the system in the manifold coordinate representation, $h$.  Shown in Fig.~\ref{fig:Latent_KSE_Forecasting-a}, is an example trajectory from the KSE.  Fig.~\ref{fig:Latent_KSE_Forecasting-b} is the $\mathcal{E}_h$ encoded manifold representation of the same trajectory. From a single encoded initial condition in the ambient space, we can perform the entire systems forecast in the manifold space. This forecasted trajectory, {for the same initial condition used to generate Fig.~\ref{fig:Latent_KSE_Forecasting-a},} is shown in Fig.~\ref{fig:Latent_KSE_Forecasting-d}. Naturally, the ambient representation of this trajectory can be completely recovered via $\mathcal{D}_h$, shown in Fig.~\ref{fig:Latent_KSE_Forecasting-c}. {Comparison of the top and bottom rows shows that the time-evolution prediction in the manifold coordinate system is quantitatively accurate for nearly 50 time units.}    We emphasize that because the KSE is a chaotic dynamical system, the ground truth and forecast will eventually diverge. Nevertheless, the relevant time scale (the Lyapunov time) of this system is $\sim 20$ time units and we achieve quantitative agreement for about two Lyapunov times. From this result, we highlight that our learned manifold coordinate system is conducive for forecasting, and our mapping functions produce good ambient space reconstruction. 

\begin{figure}
	\begin{center}
		\includegraphics[width=0.95\textwidth]{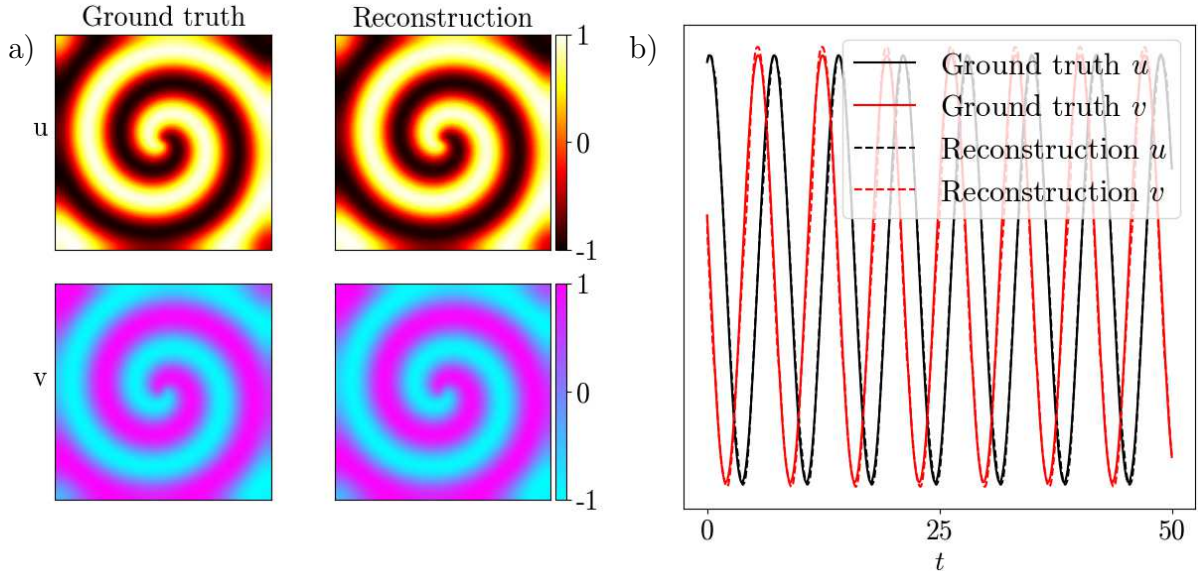}
        \captionsetup[subfigure]{labelformat=empty}
        \begin{picture}(0,0)
    	\put(-460,200){\contour{white}{ \textcolor{black}{a)}}}
    	\put(-225,200){\contour{white}{ \textcolor{black}{b)}}}
    	\end{picture} 
    	\begin{subfigure}[b]{0\textwidth}\caption{}\vspace{-10mm}\label{fig:ReactionDiffusionReconstruction_a}\end{subfigure}
     	\begin{subfigure}[b]{0\textwidth}\caption{}\vspace{-10mm}\label{fig:ReactionDiffusionReconstruction_b}\end{subfigure}
		\caption{a) Ground truth and prediction of a spiral wave produced by the lambda-omega reaction-diffusion system after one period and b) the time series ground truth and forecast of the $u$ and $v$ components at $x = 1, y =45$. Ground truth and predictions were evolved from identical initial conditions.}
		\label{fig:ReactionDiffusionReconstruction}
	\end{center}
\end{figure}

We further demonstrate the ability of IRMAE-WD to produce low-dimensional dynamical models for high-dimensional ambient systems using the lambda-omega reaction-diffusion system. We train a neural ODE, $\dot{h}=g(h;\theta_g)$, to model the dynamics of the system in the manifold coordinates, $h$, as done by \citet{Linot2021}. Using a single encoded initial position in the ambient space, we can forecast the evolution of the full ambient system in the manifold space. The ground truth of the spiral wave after one period is shown in Fig.~\ref{fig:ReactionDiffusionReconstruction_a}, along with the corresponding ambient space reconstruction decoded from the neural ODE forecast. The reconstructed spiral wave matches the ground truth, as the predicted evolution closely tracks the evolution of the true system dynamics. In Fig.~\ref{fig:ReactionDiffusionReconstruction_b}, we present the time series of the ground truth $u$ and $v$ components at a single spatial location and the reconstruction from the predicted trajectory. The predicted evolution behaves nearly  identically to the ground truth, further demonstrating the quantitative accuracy of our forecasts.

\MDGrevise{Furthermore, we briefly note that the combination of IRMAE-WD and neural ODE evolution in manifold coordinates has separately been applied to prediction of microstructural evolution in a flowing complex fluid \cite{Young.2023.10.1007/s00397-023-01412-0}.  In that case, synthetic X-ray scattering pattern data for a suspension of Brownian rigid rods in complex flows was reduced to a five-dimensional manifold and the evolution of those dimensions learned, with excellent reconstruction. }

\MDGrevise{To conclude the discussion of modeling of time evolution in the manifold coordinates, we address the issue of whether ``end-to-end" modeling, which would simultaneously determine the manifold dimension, coordinates, and evolution equation from time series data, is feasible or practical. While approaches along these lines have recently been proposed \cite{Kicic.2023.10.1016/j.cma.2023.116204}, they have not been applied to cases with dynamics more complex than limit cycles.  For systems with a high ambient dimension and complex chaotic dynamics, trajectories need to visit each region of the invariant manifold several times so that the true shape of the manifold can be ascertained. I.e. it is necessary to see the global ``shape" of the manifold before one can find a coordinate representation of it. 
All methods that we know of for estimating manifold dimension share this feature. Using ensembles of trajectories starting from different parts of state space it may be possible to circumvent this issue (cf.~\cite{Graham:1996uy}); such an approach is beyond the scope of the present work.}

\subsection{The Dynamics of Low-Rank Representation Learning}
\label{sec:dynamics}

We now turn our attention towards understanding the automatic learning of an approximately minimal representation. We glean insights by framing our network as a dynamical system, where ``space'' corresponds to layer depth in the network and ``time'' corresponds to training epoch/iteration.  In this manner, we elucidate ``when'' and ``where'' low rank behavior appears in our network.

More precisely, we will compute and track the singular value spectra for a range of intermediate latent representations, weight matrices, and update gradients as a function of \KZrevise{model layer and epoch}. We will use these spectra to estimate the rank {(based on the position of a substantial gap in the singular value spectrum of the matrix under investigation)}. The following analyses are performed on a framework with $n=4, \lambda=10^{-6}$, trained on the KSE $L=22$ dataset, which has a {64-}dimensional ambient space with a nonlinear invariant manifold with $d_m=8$. 

We first define several key weights, $W_j$, and representations, $z_j$, in the model from input to output, where $j$ is a placeholder for the position in the network. Starting from the encoder, we define the nonlinearly-activated representation immediately output from the \emph{nonlinear} portion of the encoder, $\mathcal{E}_N$, 
as $z_{\mathcal{E}N}=\mathcal{E}_N(u)$. This representation is then mapped to $\mathbb{R}^{d_z}$ by a linear layer $W_\mathcal{E}$ to result in representation $z_\mathcal{E}=\mathcal{E}(u)=W_\mathcal{E}\mathcal{E}_N(u)$. From here, the representation passes through $n$ square linear layers: $W_1,...,W_n$. The representation output after each of these layers is then $z_1,...,z_n$. Note that $z_n$ is equal to $z$ in the nomenclature of the previous sections. Finally, before arriving at the nonlinear decoder, $\mathcal{D}_N$, $z_n$ is mapped via $W_\mathcal{D}$ to the proper size: $\mathcal{D}(z_n)=\mathcal{D}_N(W_\mathcal{D} z_n)$. To summarize, a fully encoded and decoded snapshot of data is $\tilde{u}=\mathcal{D}(\mathcal{W}(\mathcal{E}(u)))=\mathcal{D}_N(W_\mathcal{D} W_n ... W_1 W_\mathcal{E}\mathcal{E}_N(u))$.

\begin{figure}
	\begin{center}
		\includegraphics[width=0.95\textwidth]{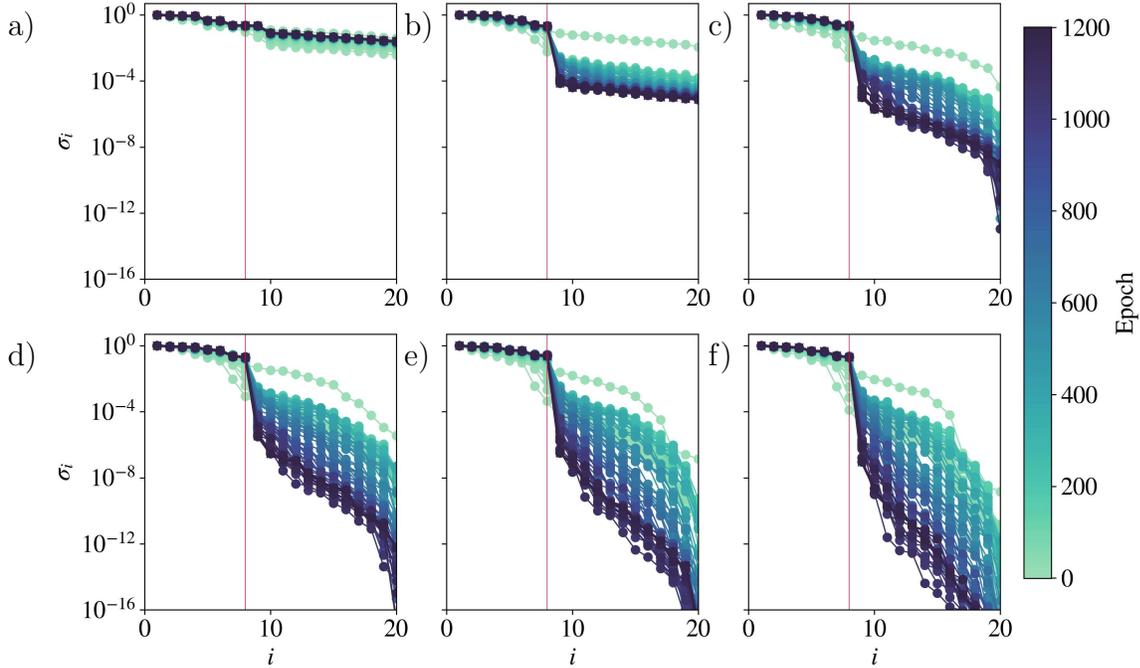}
        \captionsetup[subfigure]{labelformat=empty}
        \begin{picture}(0,0)
    	\put(-450,250){\contour{white}{ \textcolor{black}{a)}}}
    	\put(-300,250){\contour{white}{ \textcolor{black}{b)}}}
        \put(-185,250){\contour{white}{ \textcolor{black}{c)}}}
    	\put(-450,125){\contour{white}{ \textcolor{black}{d)}}}
        \put(-300,125){\contour{white}{ \textcolor{black}{e)}}}
    	\put(-185,125){\contour{white}{ \textcolor{black}{f)}}}
    	\end{picture} 
        \begin{subfigure}[b]{0\textwidth}\caption{}\vspace{-10mm}\label{fig:track_rank_representation-a}\end{subfigure}
     	\begin{subfigure}[b]{0\textwidth}\caption{}\vspace{-10mm}\label{fig:track_rank_representation-b}\end{subfigure}
        \begin{subfigure}[b]{0\textwidth}\caption{}\vspace{-10mm}\label{fig:track_rank_representation-c}\end{subfigure}
     	\begin{subfigure}[b]{0\textwidth}\caption{}\vspace{-10mm}\label{fig:track_rank_representation-d}\end{subfigure}
        \begin{subfigure}[b]{0\textwidth}\caption{}\vspace{-10mm}\label{fig:track_rank_representation-e}\end{subfigure}
     	\begin{subfigure}[b]{0\textwidth}\caption{}\vspace{-10mm}\label{fig:track_rank_representation-f}\end{subfigure}
		\caption{``Space-Time'' tracking of the singular spectra of the covariance of the representation of the data, $z_j$, trained on the KSE $L=22$ dataset: a) $z_{\mathcal{E}N}$ b) $z_{\mathcal{E}}$ c) $z_{1}$, d) $z_{2}$, e) $z_{3}$, and f) $z_{4}$ (i.e.~$z$) as a function of training epoch. Note that the spectra for a) and b) are truncated for clarity. The $d_m$ of the dataset is denoted by a vertical red line.}
		\label{fig:track_rank_representation}
	\end{center}
\end{figure}

We first perform space-time tracking of the rank of the latent representation, shown in Fig.~\ref{fig:track_rank_representation}, by computing the singular spectrum of the covariance of the data representation, $z_j$, at various intermediate layers of the network and various epochs during training. As we traverse our model in space (layer), we find that the nonlinear encoder produces a full-rank representation and is not directly responsible for transforming the data into its low-rank form, shown in Fig.~\ref{fig:track_rank_representation-a}. However, we observe that as the data progresses from the nonlinear encoder and through the non-square linear mapping to $W_1$, the learned representation is weakly low-rank, shown in Fig.~\ref{fig:track_rank_representation-b}. As the data progresses through each of the square linear blocks $W_1,...,W_n$, we observe that the unnecessary singular values/directions of the representation are further attenuated (equivalently the most essential representation directions are amplified), transforming the latent representation towards a true minimal-rank representation.

As we traverse our model in time (i.e.~epoch), we observe that the rank of the learned representation for $z_{\mathcal{E}N}$ is stagnant. In contrast, we observe for each of the sequential linear layers the rank of the representation begin as essentially full-rank, but then collectively decay into low-rank representations. We note that the representation during the early epochs ``over-correct'' to a representation that is too low of a rank to accurately capture the data, but the network automatically resolves this as training progresses.

\begin{figure}
	\begin{center}
		\includegraphics[width=0.95\textwidth]{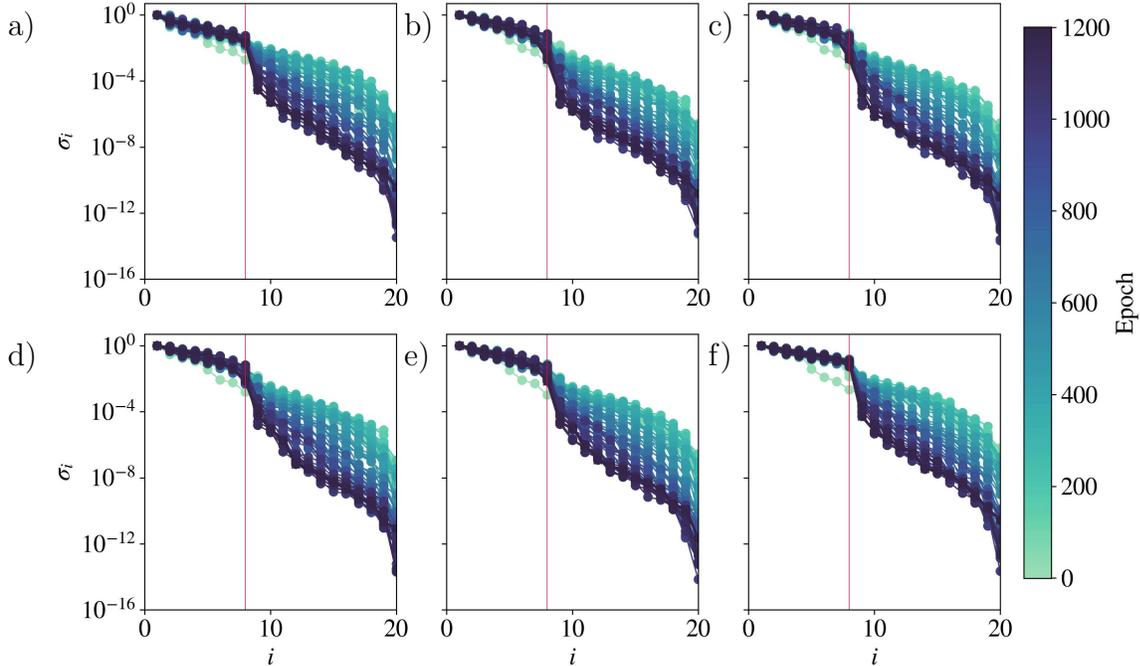}
        \captionsetup[subfigure]{labelformat=empty}
        \begin{picture}(0,0)
    	\put(-450,250){\contour{white}{ \textcolor{black}{a)}}}
    	\put(-300,250){\contour{white}{ \textcolor{black}{b)}}}
        \put(-185,250){\contour{white}{ \textcolor{black}{c)}}}
    	\put(-450,125){\contour{white}{ \textcolor{black}{d)}}}
        \put(-300,125){\contour{white}{ \textcolor{black}{e)}}}
    	\put(-185,125){\contour{white}{ \textcolor{black}{f)}}}
    	\end{picture} 
        \begin{subfigure}[b]{0\textwidth}\caption{}\vspace{-10mm}\label{fig:track_rank_gradients-a}\end{subfigure}
     	\begin{subfigure}[b]{0\textwidth}\caption{}\vspace{-10mm}\label{fig:track_rank_gradients-b}\end{subfigure}
        \begin{subfigure}[b]{0\textwidth}\caption{}\vspace{-10mm}\label{fig:track_rank_gradients-c}\end{subfigure}
     	\begin{subfigure}[b]{0\textwidth}\caption{}\vspace{-10mm}\label{fig:track_rank_gradients-d}\end{subfigure}
        \begin{subfigure}[b]{0\textwidth}\caption{}\vspace{-10mm}\label{fig:track_rank_gradients-e}\end{subfigure}
     	\begin{subfigure}[b]{0\textwidth}\caption{}\vspace{-10mm}\label{fig:track_rank_gradients-f}\end{subfigure}
		\caption{``Space-Time'' tracking of the singular spectra of the update gradient, $\mathcal{J}_j$, for a model trained on the KSE $L=22$ dataset for the a) $\mathcal{J}_\mathcal{E}$ b) $\mathcal{J}_1$, c) $\mathcal{J}_2$, d) $\mathcal{J}_3$, e) $\mathcal{J}_4$, and f) $\mathcal{J}_\mathcal{D}$ as a function of training epoch. Note that the spectra for a) and f) are truncated for clarity. The $d_m$ of the dataset is denoted by a vertical red line.}
		\label{fig:track_rank_gradients}
	\end{center}
\end{figure}

To further understand what is happening, we now perform a similar space-time analysis of our model to track the rank of the gradient updates of the weights at each layer, $\mathcal{J}_j=\nabla_{W_j}\mathcal{L}$, shown in Fig.~\ref{fig:track_rank_gradients}. Here we follow the same layer indexing convention described above. We observe in Fig.~\ref{fig:track_rank_gradients} that in early training the sequential linear layers begin with update gradients that adjust all directions in each of the latent representations. As training progresses, the singular values of the update gradients begin to decay in unnecessary directions, shifting the latent space towards a low-rank representation. Once this is achieved, the gradient updates are essentially only updating in the significant directions needed for reconstruction. From the analysis of the gradient updates, we can conclude that the framework collectively adjusts all linear layers.

\begin{figure}
	\begin{center}
		\includegraphics[width=0.95\textwidth]{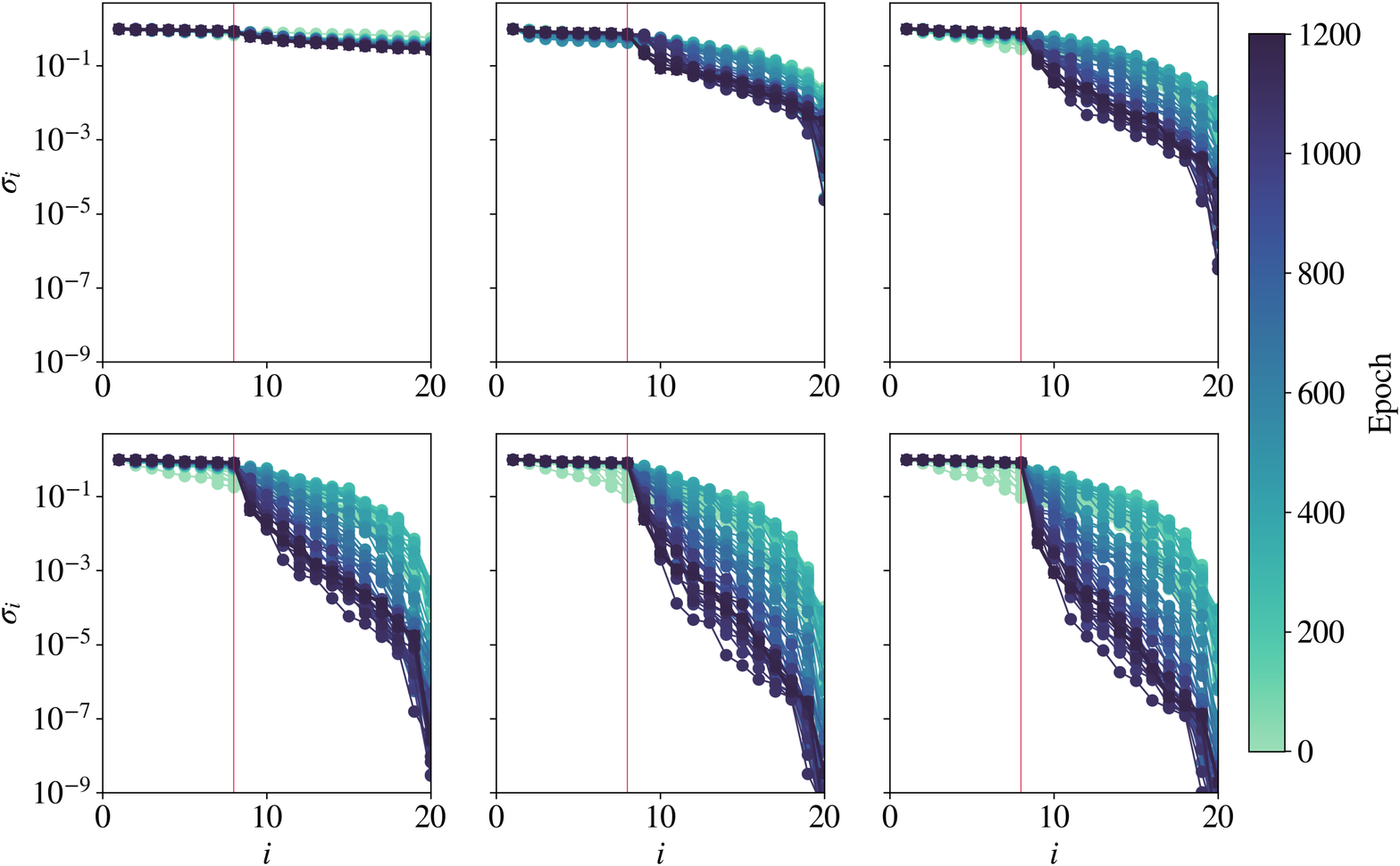}
        \captionsetup[subfigure]{labelformat=empty}
        \begin{picture}(0,0)
    	\put(-450,250){\contour{white}{ \textcolor{black}{a)}}}
    	\put(-300,250){\contour{white}{ \textcolor{black}{b)}}}
        \put(-185,250){\contour{white}{ \textcolor{black}{c)}}}
    	\put(-450,125){\contour{white}{ \textcolor{black}{d)}}}
        \put(-300,125){\contour{white}{ \textcolor{black}{e)}}}
    	\put(-185,125){\contour{white}{ \textcolor{black}{f)}}}
    	\end{picture} 
        \begin{subfigure}[b]{0\textwidth}\caption{}\vspace{-10mm}\label{fig:track_rank_weights_compounded-a}\end{subfigure}
     	\begin{subfigure}[b]{0\textwidth}\caption{}\vspace{-10mm}\label{fig:track_rank_weights_compounded-b}\end{subfigure}
        \begin{subfigure}[b]{0\textwidth}\caption{}\vspace{-10mm}\label{fig:track_rank_weights_compounded-c}\end{subfigure}
     	\begin{subfigure}[b]{0\textwidth}\caption{}\vspace{-10mm}\label{fig:track_rank_weights_compounded-d}\end{subfigure}
        \begin{subfigure}[b]{0\textwidth}\caption{}\vspace{-10mm}\label{fig:track_rank_weights_compounded-e}\end{subfigure}
     	\begin{subfigure}[b]{0\textwidth}\caption{}\vspace{-10mm}\label{fig:track_rank_weights_compounded-f}\end{subfigure}
		\caption{``Space-Time'' tracking of the singular spectra of the effective linear layer, $W_{j,\eff}$, for a model trained on the KSE $L=22$ dataset. The singular spectra for the effective weight matrix a) $W_{\mathcal{E},\eff}$ b) $W_{1,\eff}$, c) $W_{2,\eff}$, d) $W_{3,\eff}$, e) $W_{4,\eff}$, and f) $W_{\mathcal{D},\eff}$ as a function of training epoch. Note that the spectra for a) and f) are truncated for clarity. The $d_m$ of the dataset is denoted by a vertical red line.}
		\label{fig:track_rank_weights_compounded}
	\end{center}
\end{figure}

As linear layers in sequence can be subsumed into a single linear layer by simply computing the product of the sequence, we also investigate the rank of the \emph{effective} layer weight matrix itself, $W_{j,\eff}$ (e.g.\ $W_{2,\eff}=W_2 W_1 W_\mathcal{E}$) in space and time, shown in  Fig.~\ref{fig:track_rank_weights_compounded}. We show in Fig.~\ref{fig:track_rank_weights_compounded} as the linear layers compound deeper into the network, the effective rank of the layers approaches $d_m$. This coincides with the observation made in Fig.~\ref{fig:track_rank_representation}. We conclude here that the sequential linear layers work together to form an effective  rank $d_m$ weight matrix, projecting the data onto a space of dimension $d_m$. We highlight here that while the network automatically learns a linearly separated $d_m$ representation, the manifold of the original dataset is nonlinear in nature and is nonlinearly embedded in the ambient space--this feature is captured by the nonlinear encoding and decoding blocks. \KZrevise{Finally, we comment that when weight-sharing is implemented across the linear blocks $W_j$ (i.e.\ $W_j$ are equal) we lose regularization as weight-sharing decreases the effective number of linear layers}.


\begin{figure}
	\begin{center}
		\includegraphics[width=0.9\textwidth]{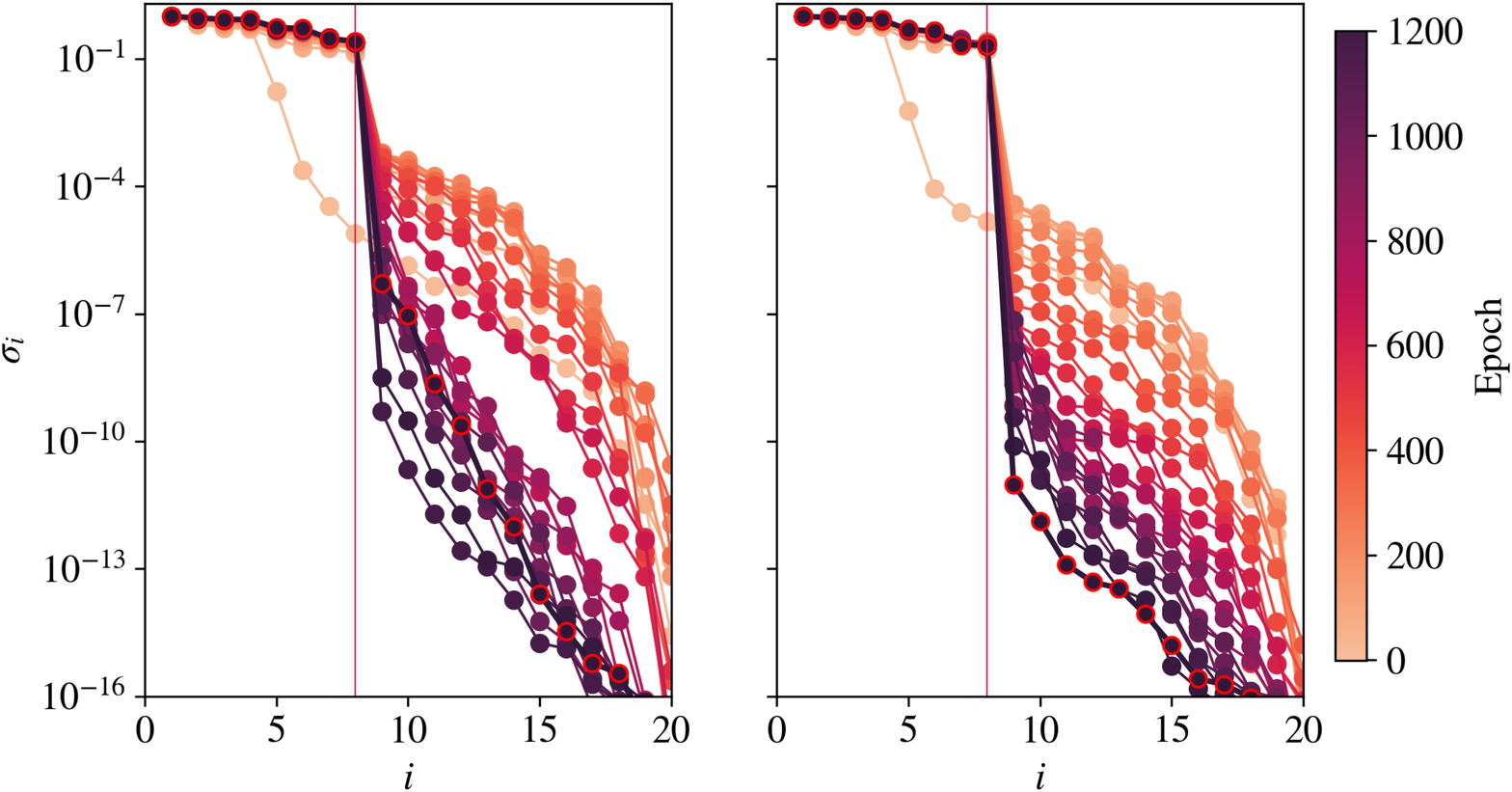}
        \captionsetup[subfigure]{labelformat=empty}
        \begin{picture}(0,0)
    	\put(-420,200){\contour{white}{ \textcolor{black}{a)}}}
    	\put(-230,200){\contour{white}{ \textcolor{black}{b)}}}
    	\end{picture} 
    	\begin{subfigure}[b]{0\textwidth}\caption{}\vspace{-10mm}\label{fig:KSE_track_SV-a}\end{subfigure}
     	\begin{subfigure}[b]{0\textwidth}\caption{}\vspace{-10mm}\label{fig:KSE_track_SV-b}\end{subfigure}
		\caption{``Space-Time'' tracking of the singular spectra of the covariance of the representation of the data, $z$, trained on the KSE $L=22$ dataset: a) an AE with only implicit regularization (IRMAE) b) an AE with implicit regularization and weight-decay (IRMAE-WD). The $d_m$ of the dataset is denoted by a vertical red line and the final learned latent spectrum is outlined in red markers.}
		\label{fig:KSE_track_SV}
	\end{center}
\end{figure}

\KZrevise{We conclude this section with a comparison between our proposed framework IRMAE-WD, which utilizes implicit regularization and weight-decay, and one that only utilizes implicit regularization, IRMAE. We show in Fig. \ref{fig:KSE_track_SV} the learning dynamics of the data covariance of the latent representation for each. Fig. \ref{fig:KSE_track_SV-a} shows the dynamics in the absence of weight decay where we observe that the trailing singular values first drift upward in the first 100 epochs, followed by decay and then growth again as training proceeds. The addition of weight decay, as shown in Fig. \ref{fig:KSE_track_SV-b}, leads to monotonic decay of the trailing singular values.} {These observations are consistent with the linear IRMAE-WD analysis in Appendix \ref{sec:AppendixC}, which, in the absence of weight-decay predicts directions with eigenvalues at zero in which the training dynamics will drift. Adding weight decay makes these eigenvalues negative, aiding convergence.}

\section{Conclusions} \label{sec:Conclusions}

In this paper, we build upon observations made by \citet{Jing2020} and present an autoencoder framework, denoted IRMAE-WD, that combines implicit regularization {with internal linear layers} and weight decay to automatically estimate the underlying dimensional $d_m$ of the manifold on which the data lies. This framework simultaneously learns an ordered and orthogonal manifold coordinate representation as well as the mapping functions between the ambient space and manifold space, allowing for out-of-sampling projections. Unlike other autoencoder methods, we accomplish this without parametric model sweeps or relying on secondary algorithms, requiring only that the bottleneck dimension $d_z$ of the autoencoder satisfies  $d_z>d_m$.
 
We demonstrated our framework by estimating the manifold dimension for a series of finite and (discretized) infinite-dimensional systems that possess linear and nonlinear manifolds. We show that it outperforms several state-of-the-art estimators for systems with nonlinear embedded manifolds and is even accurate for \KZrevise{relatively large manifold dimensions, $d_m\approx 40$. However, the ambient dimensions of our test systems are still small relative to the demands of many industrially relevant applications, such as turbulent fluid flows where the ambient dimension $d_u$ (number of Fourier modes or grid points) can easily exceed $10^6$ and $d_m$ is suspected to increase very strongly with Reynolds number (flow strength). We aim with future work to efficiently extend IRMAE-WD  to these high-dimensional systems.}

We demonstrate that our framework can be naturally extended for applications of state-space modeling and forecasting with the Kuramoto-Sivashinsky equation \KZr{and the lambda-omega reaction-diffusion system}. Using a neural ODE, we learned the dynamics of 
\KZr{these datasets} in the manifold representation and showed that the ambient space representation can be accurately recovered at any desired point in time.
 
Our analyses of the training process in ``space'' (layer)  and ``time'' (epoch) indicate that low-rank learning appears simultaneously in all linear layers.  We highlight that the nonlinear encoder is not directly responsible for learning a low-rank representation, but rather each of the sequential linear layers work together by compounding the approximately low-rank features in the latent space, effectively amplifying the relevant manifold directions and equivalently attenuating superfluous modes. {Analysis of a linear autoencoder with the IRMAE-WD architecture illustrates the role of the linear layers in accelerating collective convergence of the encoder, decoder, and internal layers as well as the role of weight-decay in breaking degeneracies that limit convergence in its absence.} {On the theoretical side,  while the linear autoencoder analysis presented in Appendix \ref{sec:AppendixC} provides some insight, it is quite limited, and further, more sophisticated studies are necessary to better understand the method\MDGrevise{, even in the linear, much less} the fully nonlinear setting.}
 
Finally, we demonstrate that our framework is quite robust to choices of $L_2$ regularization (weight decay) parameter $\lambda$ and number of linear layers $n$. We show that in a large envelope of regularization parameters we achieve accurate estimations of $d_m$ without sacrificing accuracy (MSE). We also show that $\lambda$ can help reduce the contribution of superfluous singular directions in the learned latent space.

\KZrevise{While the present work is motivated by complex \textit{deterministic} dynamical systems, we acknowledge that many practical systems of interest are stochastic or noisy and the data may only lie \textit{near}, but not precisely on a finite-dimensional manifold and we aim to robustly extend IRMAE-WD to these systems in future work.}



\section*{Acknowledgments}
This work was supported by Office of Naval Research grant N00014-18-1-2865 (Vannevar Bush Faculty Fellowship). We also wish to thank the UW-Madison College of Engineering Graduate Engineering Research Scholars (GERS) program and acknowledge funding through the Advanced Opportunity Fellowship (AOF) as well as the PPG Fellowship.

\MDGrevise{\section*{Data availability statement}
Code and sample data that support the findings of this study are openly available at \url{https://github.com/mdgrahamwisc/IRMAE_WD} 
}

\appendix

\section{Model Architecture and Parameters} \label{sec:AppendixA}
\begin{table}[ht]
\caption{Here we list the architecture and parameters utilized in the studies of this paper. For brevity, the decoders, $\mathcal{D}$, of each architecture is simply mirrors of the encoder, $\mathcal{E}$ with activations ReLU/ReLU/lin. Each network has $n$ sequential linear layers with shape $d_z\times d_z$ between the encoder and decoder. Learning rates were set to $10^{-3}$ and with mini-batches of 128.} 
\centering 
\begin{tabular}{c c c c c c} 
\hline\hline 
Dataset & $\mathcal{E}$ & Activation & $d_z$ & $n$ & $\lambda$\\ [0.5ex] 
\hline 
5D Noise & 20/128/64/20 & ReLU/ReLU/lin & 20 & 4 & $10^{-2}$\\
Arch. Lorenz & 4/128/64/4 & ReLU/ReLU/lin & 4 & 4 & $10^{-6}$\\
2Torus & 3/256/128/10 & ReLU/ReLU/lin & 10 & 4 & $10^{-2}$\\
KSE $L=22$ & 64/512/256/20 & ReLU/ReLU/lin & 20 & 4& $10^{-6}$\\
KSE $L=44$ & 64/512/256/30 & ReLU/ReLU/lin & 30 & 4& $10^{-6}$\\
KSE $L=66$ & 64/512/256/50 & ReLU/ReLU/lin & 50 & 4& $10^{-6}$\\
KSE $L=88$ & 20/512/256/80 & ReLU/ReLU/lin & 80 & 4 & $10^{-6}$\\ [1ex] 
\hline 
\end{tabular}
\label{table:arch} 
\end{table}

\section{Application to the MNIST Handwriting Dataset} \label{sec:AppendixB}

Here we apply IRMAE-WD to the MNIST dataset and compare to \citet{Jing2020}. We utilize the same convolutional autoencoder architecture parameters that they used, with the following parameters and architecture: $4\times4$ kernel size with stride 2, padding 1, a learning rate of $10^{-3}$, and $\lambda=10^{-6}$. Here $\text{Conv}$, $\text{ConvT}$, and $\text{FC}$ correspond to a convolutional layer, transposed-convolutional layer, and fully connected (not activated) layer, respectively.

\begin{figure}
	\begin{center}
		\includegraphics[width=0.90\textwidth]{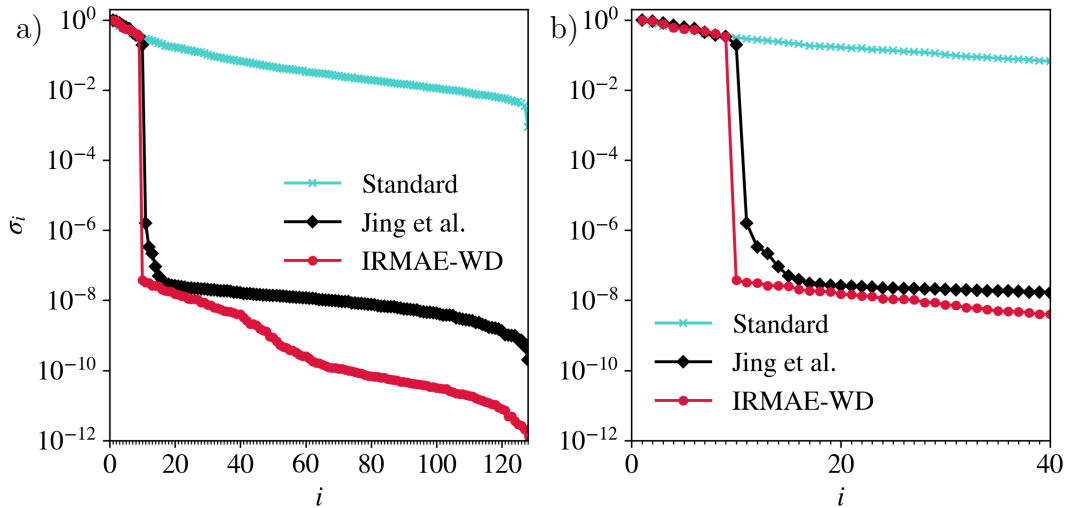}
        \captionsetup[subfigure]{labelformat=empty}
        \begin{picture}(0,0)
    	\put(-415,190){\contour{white}{ \textcolor{black}{a)}}}
    	\put(-213,190){\contour{white}{ \textcolor{black}{b)}}}
    	\end{picture} 
    	\begin{subfigure}[b]{0\textwidth}\caption{}\vspace{-10mm}\label{fig:SV_MNIST-a}\end{subfigure}
     	\begin{subfigure}[b]{0\textwidth}\caption{}\vspace{-10mm}\label{fig:SV_MNIST-b}\end{subfigure}
		\caption{Singular value spectra obtained from models trained over the MNIST handwriting dataset a) full spectra and b) zoomed in spectra.}
		\label{fig:SV_MNIST}
	\end{center}
\end{figure}

In Fig. \ref{fig:SV_MNIST} we show that IRMAE-WD, which utilizes both implicit and weight regularization, learns a $d_m=9$ representation for the MNIST handwriting dataset while \citet{Jing2020}, which only utilizes implicit regularization, learns a $d_m=10$ representation. We further highlight that the trailing singular values from our model sharply decays several orders of magnitude lower than the \citet{Jing2020} model. We finally note that the latent space from the \citet{Jing2020} model exhibits a broader tail, especially near the significant singular values.  \KZrevise{We find that despite our model utilizing one fewer degree of freedom to model the MNIST data, it produces an MSE that is comparable to \citet{Jing2020} when trained using their parameters ($1.0\cdot 10^{-2}$ vs. $9.5\cdot 10^{-3}$)}.

\begin{table}[ht]
\caption{Convolutional autoencoder and architecture for MNIST Handwriting Dataset} 
\centering 
\begin{tabular}{c | c} 
\hline\hline 
Encoder & Decoder\\ [0.5ex] 
\hline 

$x\in\mathbb{R}^{32\times32\times1}$ & $z\in\mathbb{R}^{128}$\\
$\rightarrow \text{Conv}_{32} \rightarrow \text{ReLU}$ & $\rightarrow \text{FC}_{4096}$ \\
$\rightarrow \text{Conv}_{64} \rightarrow \text{ReLU}$ & $\rightarrow \text{reshape}_{8\times8\times64}$ \\
$\rightarrow \text{Conv}_{128} \rightarrow \text{ReLU}$ & $\rightarrow \text{ConvT}_{64} \rightarrow \text{ReLU}$ \\
$\rightarrow \text{Conv}_{256} \rightarrow \text{ReLU}$ & $\rightarrow \text{ConvT}_{32} \rightarrow \text{ReLU}$\\
$\rightarrow \text{flatten}_{1024}$ & $\rightarrow \text{ConvT}_{1} \rightarrow \text{Tanh}$\\
$\rightarrow \text{LC}_{128} \rightarrow z\in\mathbb{R}^{128}$ & $\rightarrow \hat{x}\in\mathbb{R}^{32\times32\times1}$\\ [1ex] 
\hline 
\end{tabular}
\label{table:archMNIST} 
\end{table}

\section{Analysis of linear autoencoders with internal linear layers and weight decay} \label{sec:AppendixC}


\subsection{Formulation}

To gain some insight into the performance of autoencoders with additional linear layers and weight decay, we present here an analysis of gradient descent for an idealized case of a linear autoencoder with  one or more internal linear layers. We begin with the formalism with a single internal linear layer. The input is denoted $u\in \reals^{d_u}$, encoder $E\in\reals^{{d_z}\times {d_u}}$, decoder $D\in\reals^{{d_u}\times {d_z}}$, internal linear layer $W\in\reals^{{d_z}\times {d_z}}$ and output $\xt=DWE u\in\reals^{{d_u}}$. We define the latent variable preceding the linear layer as $h=Eu\in\reals^{d_z}$ and the one following it as $w=Wh=WEu\in\reals^{d_z}$.  For a conventional autoencoder, $W=I^{d_zd_z}$, where the notation $I^{mm}$ denotes the $m\times m$ identity matrix.
We will consider the simple loss function
\begin{equation*}
\loss	=\langle||\xt-u||^{2}_{2}\rangle+\lambda_E(||E||^2_F+||D||^2_F)+\lambda_W||W||^2_F,
\end{equation*} 
where $\langle\cdot\rangle$ denotes ensemble average (expected value).  
First the converged equilibrium solution of the minimization problem for the loss will be considered, and then the convergence of the solution to the minimum.

We are particularly interested in the case where the data lies on an $r$-dimensional subspace of $\reals^{d_u}$, or equivalently $\rank\langle u u^T\rangle =r$, and we assume that the dimension $m$ of the hidden layers is chosen so that $m>r$.   

We can write the loss as 
\begin{equation*}
	\loss=\langle u^T(E^TW^TD^TDWE-(DWE+E^TW^TD^T))u\rangle+\langle u^Tu\rangle+\lambda_E( \trace{EE^T}+\trace{DD^T})+\lambda_W \trace{WW^T}.
\end{equation*}
In index notation we can write 
\begin{align*}
	\loss&=\langle u_k E^T_{kl}W^T_{lm}D^T_{mn}D_{no}W_{op}E_{pq} u_q\rangle\\
	&-\langle u_k(D_{kl}W_{lm}E_{mn}+E^T_{kl}W^T_{lm}D^T_{mn})u_n\rangle\\
	&+\langle u_k u_k\rangle+\lambda_E (E_{kl}E_{kl}+D_{kl}D_{kl})+\lambda_W W_{kl}W_{kl}.
\end{align*}
Taking partial derivatives yields
\begin{align*}
	\frac{\partial \loss}{\partial E_{ij}}&=\langle 2 W^T_{im}D^T_{mn}(DWE-I)_{no})u_ou_j\rangle+2\lambda_E E_{ij}=\langle 2 W^T_{im}D^T_{mn}(\xt_n-u_n)u_j\rangle+2\lambda_E E_{ij},\\
	\frac{\partial \loss}{\partial W_{ij}}&=\langle 2D^T_{ik}(DWE-I)_{kl}u_lE_{jm}u_m\rangle+2\lambda W_{ij}=\langle 2D^T_{ik}(\xt_k-u_k)h_j\rangle+2\lambda_E W_{ij},\\
	\frac{\partial \loss}{\partial D_{ij}}&=\langle 2(DWE-I)_{ik}u_k(WE)_{jl}u_l\rangle+2\lambda_E D_{ij}=2(\xt_i-u_i)w_j+2\lambda_E D_{ij}.
\end{align*}
We consider a highly idealized dataset where $\langle u u^T\rangle=\sigma^2 I^{rd_ud_u}$, where $I^{rpq}$ is an $p\times q$ matrix (with $p,q>r$) whose first $r$ diagonal elements are unity and all others are zero. 
Now
\begin{align*}
	\frac{\partial \loss}{\partial E_{ij}}&= 2\sigma^2 W^T_{im}D^T_{mn}(DWE-I)_{no})\Irnn_{oj}+2\lambda_E E_{ij},\\
	\frac{\partial \loss}{\partial W_{ij}}&=2\sigma^2 D^T_{ik}(DWE-I)_{kl}E_{jm}\Irnn_{lm}+2\lambda_W W_{ij},\\
	\frac{\partial \loss}{\partial D_{ij}}&= 2\sigma^2(DWE-I)_{ik}(WE)_{jl}\Irnn_{kl}+2\lambda_E D_{ij}.
\end{align*}
In matrix-vector notation this becomes 
\begin{align*}
	\frac{\partial \loss}{\partial E}&= 2\sigma^2 W^TD^T(DWE-I)\Irnn+2\lambda_E E,\\
	\frac{\partial \loss}{\partial W}&=2\sigma^2 D^T(DWE-I)\Irnn E^T+2\lambda_W W,\\
	\frac{\partial \loss}{\partial D}&= 2\sigma^2(DWE-I)\Irnn(WE)^T+2\lambda_E D.
\end{align*}

\subsection{Equilibrium solutions}
At convergence, these derivatives vanish.
For the moment, we set $\lambda_E=0$.  We first consider the solution in absence of the internal linear layer: i.e. when $W=I^{d_zd_z}$.  Now $\frac{\partial \loss}{\partial E}$ and $  \frac{\partial \loss}{\partial D}$ will vanish when
\begin{equation*}
	(DE-I)\Irnn=DE \Irnn-\Irnn=0.
\end{equation*}
This has ``full rank" solution $E=D^T=I^{d_zd_u}$, which satisfies $DE-I=0$, as well as ``rank $r$" solution $E=D^T=\Irmn$. This does not satisfy $DE-I=0$, but does satisfy $DE \Irnn-\Irnn=0$.  
If we include a nontrivial linear layer $W$, we than have 
\begin{equation*}
	(DWE-I)\Irnn=DWE \Irnn-\Irnn=0.
\end{equation*}
it is clear that the rank $r$ solution $E=D^T=\Irnm$, along with the rank $r$ choice $W=\Irmm$ continues to be a solution, as does the full rank solution with $E=D^T=I^{d_zd_u}$ with $W=I^{d_zd_z}$.  

In the presence of weight decay the situation is more complex, and we will only consider the case $\lambda_E=\lambda_W=\lambda$.  Defining a new parameter $\zeta=\lambda/\sigma^2$, and taking this parameter to be small,  a perturbation solution of the form 
\begin{equation}
E=\Irmn(1+\alpha \zeta+O(\zeta^2)), W=\Irmm(1+\beta\zeta+O(\zeta^2)), D=\Irnm(1+\gamma\zeta+O(\zeta^2))\label{eq:equil}
\end{equation} 
can be found. Plugging into the equilibrium conditions $\frac{\partial \loss}{\partial E}=0, \frac{\partial \loss}{\partial W}=0,\frac{\partial \loss}{\partial D}=0$ and neglecting terms of $O(\zeta^2)$  yields in each case 
\begin{equation*}
	\alpha+\beta+\gamma+1=0. 
\end{equation*}
Thus there is a whole family of solutions to the equilibrium problem with weight decay. For future reference we will write this solution (up to $O(\zeta)$) as 
\begin{equation}
	\begin{aligned}
E&=a\Irmn, \quad a=1+\alpha \zeta,\\
W&=b\Irmm,\quad b=1+\beta\zeta,\\
D&=c\Irnm, \quad c=1+\gamma\zeta	.
	\end{aligned}\label{eq:equil2}	
\end{equation}

\subsection{Convergence of gradient descent}

\subsubsection{Dynamic model}

Now we turn to the issue of convergence of gradient descent to an equilibrium solution.  Here we consider a very simple ordinary differential equation model of the process, where $t$ is a pseudotime representing number of gradient descent steps:
\begin{equation}
\begin{aligned}
\frac{dE}{dt}&=-\frac{1}{2\sigma^2}\frac{\partial \loss}{\partial E}= - W^TD^T(DWE-I)\Irnn-\zeta E,\\
\frac{dW}{dt}&=-\frac{1}{2\sigma^2}	\frac{\partial \loss}{\partial W}=- D^T(DWE-I)\Irnn E^T-\zeta W,\\
\frac{dD}{dt}&=-\frac{1}{2\sigma^2}	\frac{\partial \loss}{\partial D}= -(DWE-I)\Irnn(WE)^T-\zeta D.
\end{aligned}\label{eq:dynamics}
\end{equation}
This is a high-dimensional and highly nonlinear system; to make progress we consider only the dynamics near convergence, linearizing the system around the converged solution \eqref{eq:equil}.  That is, we set
\begin{equation}
	\begin{aligned}
E&=a\Irmn+\epsilon\hat{E},\\
W&=b\Irmm+\epsilon\hat{W},\\
D&=c\Irnm+\epsilon\hat{D},
	\end{aligned}	
\end{equation}
where $\hat{E},\hat{W}, \hat{D}$ are perturbations aways from the converged solution. 
Inserting these expressions into \eqref{eq:dynamics}, and neglecting terms of $O(\epsilon^2)$ yields
\begin{equation}
\begin{aligned}
\frac{d\Ehat}{dt}&=- bc\left[ab \Irmn\Dhat\Irmn+ac\Irmm\Lhat\Irmn+bc\Irnm\Ehat\Irmm\right] -\zeta \Ehat,\\
\frac{d\Lhat}{dt}&=- ab\left[ab \Irmn\Dhat\Irmm+ac\Irmm\Lhat\Irmm+bc\Irmm\Ehat\Irnm\right] -\zeta \Lhat,\\
\frac{d\Dhat}{dt}&=- ac\left[ab \Irnm\Dhat\Irmm+ac\Irnm\Lhat\Irmm+bc\Irnm\Ehat\Irnm\right] -\zeta \Dhat.
\end{aligned}\label{eq:lineardynamics}
\end{equation}
We can now make some important general statements about the solutions.  First, observe that the terms in the square brackets will always yield matrices for which only the upper left $r\times r$ block is nonzero. Furthermore for any nonzero $\zeta$, all terms outside this block will be driven to zero.  Finally, observe that \eqref{eq:lineardynamics} will have time-dependent solutions of the form $\Ehat(t)=\mathcal{E}(t)\Irmn,\Lhat(t)=\mathcal{W}(t)\Irmm,\Dhat(t)=\mathcal{D}(t)\Irnm$, where $\Ecal,\Dcal$, and $\Wcal$ are \emph{scalar} functions of time. The evolution equation for these perturbations is 
\begin{equation}
\begin{aligned}
\frac{d\Ecal}{dt}&=- bc\left[ab \Dcal+ac\Wcal+bc\Ecal\right] -\zeta \Ecal,\\
\frac{d\Wcal}{dt}&=- ab\left[ab \Dcal+ac\Wcal+bc\Ecal\right] -\zeta \Wcal,\\
\frac{d\Dcal}{dt}&=- ac\left[ab \Dcal+ac\Wcal+bc\Ecal\right] -\zeta \Dcal.
\end{aligned}\label{eq:scalarlineardynamics}
\end{equation}
 Hereinafter, we will consider solutions in this invariant subspace, where a fairly complete characterization of the linearized dynamics is possible.  
 
 \subsubsection{Linear layers speed collective convergence of weights}
 
  The situation is simplest when there is no weight decay: $\zeta=0$.  Now $a=b=c=1$ and \eqref{eq:scalarlineardynamics} simplifies to 
\begin{equation}
\begin{aligned}
\frac{d\Ecal}{dt}&=- \left[\Dcal+\Wcal+\Ecal\right],\\
\frac{d\Wcal}{dt}&=- \left[\Dcal+\Wcal+\Ecal\right],\\
\frac{d\Dcal}{dt}&=- \left[\Dcal+\Wcal+\Ecal\right].
\end{aligned}\label{eq:scalarlineardynamics2}
\end{equation}
Adding these equations together yields that
\begin{equation*}
	\frac{d}{dt}(\Dcal+\Wcal+\Ecal)=-3(\Dcal+\Wcal+\Ecal).
\end{equation*} 
So the ``collective" weight (perturbation) $\mathcal{C}=\Dcal+\Wcal+\Ecal$ decays as $e^{-\rho_1 t}$ where $\rho_1=3$.  

\MDGrevise{We can also find an evolution equation for the loss. For small $\epsilon$,
\begin{equation}
    \Lcal=r\sigma^2\epsilon^2(\Dcal+\Wcal+\Ecal)^2.
\end{equation}
That is, the loss is proportional to the square of the collective weight $\Ccal$. Since $\Ccal(t)=\Ccal(0)e^{-\rho_1 t}$ we then find that 
\begin{equation}
    \Lcal(t)=\Lcal(0)e^{-2\rho_1 t}.
\end{equation}
}

More generally ,we can write \eqref{eq:scalarlineardynamics2} in matrix-vector form
\begin{equation}
\frac{d}{dt}\begin{bmatrix}\Ecal \\ \Wcal \\ \Dcal\end{bmatrix}=A
\begin{bmatrix}\Ecal \\ \Wcal \\ \Dcal\end{bmatrix},\qquad A=-\begin{bmatrix}
	1& 1 & 1 \\
	1& 1 & 1 \\
	1& 1 & 1 
\end{bmatrix}.\label{eq:scalarlineardynamics3}
\end{equation}
This has general solution
\begin{equation*}
\begin{bmatrix}\Ecal(t) \\ \Wcal(t) \\ \Dcal(t) \end{bmatrix}=C_1 e^{-3t} v_1+C_2v_2+C_3 v_3
\end{equation*}
with 
\begin{equation*}
v_1=\frac{1}{\sqrt{3}}\begin{bmatrix} 1\\ 1\\ 1 \end{bmatrix}, \qquad v_2=\frac{1}{\sqrt{2}}\begin{bmatrix} -1\\ 0\\ 1 \end{bmatrix}, \qquad v_3=\frac{1}{\sqrt{2}}\begin{bmatrix} -1\\ 1\\ 0 \end{bmatrix},
\end{equation*}
and $C_i=[\Ecal(0),\Wcal(0),\Dcal(0)]^Tv_i$.  Therefore, while the collective weight $\Dcal+\Wcal+\Ecal$ decays as $e^{-\rho_1t}$, and the loss as $e^{-2\rho_1 t}$, the quantities $\Dcal-\Ecal$ and $\Wcal-\Ecal$ do not decay at all, because of the two zero eigenvalues of the matrix $G$. This fact will limit the performance of gradient descent \MDGrevise{in generating low-rank weight matrices} in the absence of weight decay. 
(We see below that weight decay breaks the degeneracy of the dynamics.) 

Now we proceed to the question of how the number of internal linear layers affects convergence.  To consider the case of no internal linear layers, we simply set $\Lhat$ and thus $\Wcal$ to zero --- the matrix $W$ is simply fixed at the identity. Now \eqref{eq:scalarlineardynamics3} reduces to 
\begin{equation}
\frac{d}{dt}\begin{bmatrix}\Ecal \\ \Dcal\end{bmatrix}=-
\begin{bmatrix}
	1& 1 \\
	1& 1  
\end{bmatrix}
\begin{bmatrix}\Ecal \\ \Dcal\end{bmatrix}.\label{eq:scalarlineardynamics4}
\end{equation}
Now the collective weight variable $\Dcal+\Ecal$ decays as $e^{-\rho_0 t}$, with $\rho_0=2$, rather than $e^{-3t}$ when we had an internal linear layer -- this added layer accelerates convergence \MDGrevise{along the collective eigendirection}. 

What if we add additional linear layers, for a total of $n$, by replacing $W$ with a product $W_{n}W_{n-1}W_{n-2}\cdots W_{1}$? Without loss of generality we can take the converged value of each of these matrices (in the absence of weight decay) to be $\Irmm$. In considering the linearized dynamics we use the result
\begin{align*}
	W_{n}W_{n-1}W_{n-2}\cdots W_{1}&=(\Irmm+\epsilon \hat{W}_{n})(\Irmm+\epsilon \hat{W}_{n-1})(\Irmm+\epsilon \hat{W}_{n-2})\cdots (\Irmm+\epsilon \hat{W}_{1})\\
	&=\Irmm+\epsilon(\hat{W}_{n}+\hat{W}_{n-1}+\hat{W}_{n-2}+\cdots +\hat{W}_{1})+O(\epsilon^2).
\end{align*} 
Taking $\hat{W}_{i}=\Wcal_{i}\Irmm$ and following the same process as above yields the following set of equations for the linearized dynamics:
\begin{equation}
\frac{d}{dt}\begin{bmatrix}\Ecal \\ \Wcal_{n} \\ \Wcal_{n-1} \\ \vdots \\\Wcal_{1} \\ \Dcal\end{bmatrix}=-
\begin{bmatrix}
	1& 1 & 1 &\cdots & 1 & 1 \\
	1& 1 & 1 &\cdots & 1 & 1 \\
	1& 1 & 1 &\cdots & 1 & 1 \\
	\vdots & \vdots  & \vdots &  \ddots & \vdots & \vdots\\
	1& 1 & 1 &\cdots & 1 & 1 
\end{bmatrix}
\begin{bmatrix}\Ecal \\ \Wcal_{n} \\ \Wcal_{n-1} \\ \vdots \\\Wcal_{1} \\ \Dcal\end{bmatrix}\label{eq:scalarlineardynamics5}
\end{equation}
By adding these equations together we find that the collective weight  for this case $\mathcal{C}=\Dcal+\sum_{i=1}^n\Wcal_{i}+\Ecal$ decays as $e^{-\rho_n t}$, with the decay rate $\rho_n$ for an $n$ layer network given by 
\begin{equation}
	\rho_n=2+n. \label{eq:llayerdecayrate}
\end{equation}
\MDGrevise{Similarly, with additional internal layers the loss is still proportional to $\Ccal^2$, so it decays with rate $2\rho_n$. The decay rate of $\Ccal$ and $\Lcal$ relative to the case of no internal layers is then
\begin{equation}
    \frac{\rho_n}{\rho_0}=1+\frac{n}{2}.\label{eq:noWDscaling}
\end{equation}
}

The origin of this increase in convergence rate for the collective weight variable $\mathcal{C}$ (and loss) is the basic autoencoder loss structure -- for every layer, the combination $DWE-I$ appears, so the gradients for all layers will have a common structure containing the collective weight $\mathcal{C}$. The more internal linear layers, the faster this collective weight converges.  

\begin{figure}
	\begin{center}
		\includegraphics[width=0.7\textwidth]{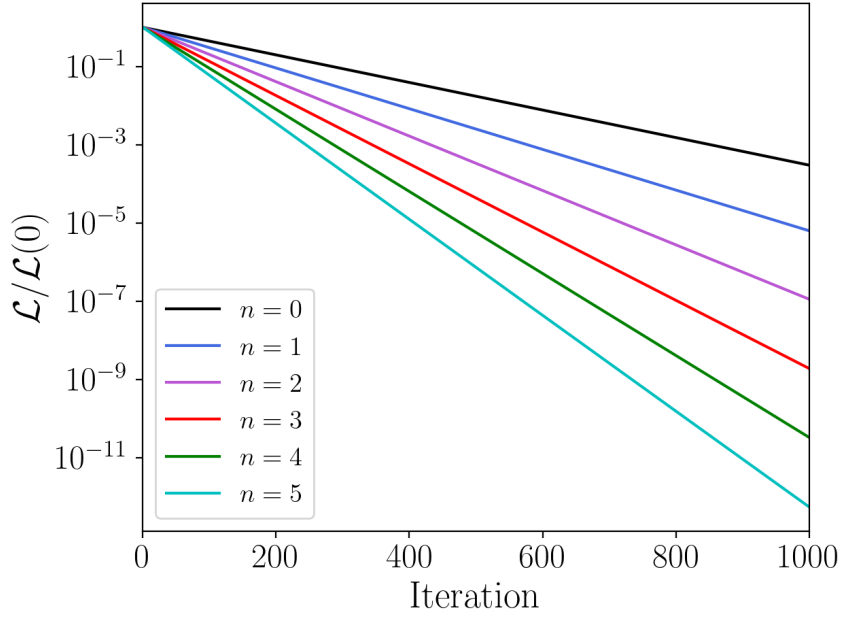}
 		\caption{Evolution of relative loss (mean squared error) for gradient descent of a linear network with diagonal perturbations.}
		\label{fig:diagnoWD}
	\end{center}
\end{figure}

\MDGrevise{To illustrate this result with an example, we considered a data set with zero mean and covariance $\sigma^2I^{r d_u d_u}$ for $r=5$ and $d_u=100$, and used internal layers of dimension $d_z=20$.  We perturbed all the diagonal elements of the weight matrices away from their equilibrium values with small zero-mean noise (the off-diagonal elements remained zero) and performed gradient descent from this initial condition.  While this is a fairly specific perturbation, it is more general than the one prescribed above (where all of the diagonal elements of each matrix would be perturbed by the same amount). The evolution of the loss, normalized by the initial value, is shown in Figure \ref{fig:diagnoWD}; the decay rates agree perfectly with the scaling of Eq.~\ref{eq:noWDscaling}.}

Now there are $n+1$ zero eigenvalues indicating directions where gradient descent does not act: $\Dcal-\Ecal$ and $\Wcal_{i}-\Ecal, i=1,\ldots n$. Thus, while addition of internal linear layer speeds convergence of the collective weight, these degenerate directions remain, limiting the overall performance of the gradient descent process.  

\subsubsection{Weight decay breaks degeneracy and leads to asymptotic stability}

Addition of weight decay complicates the analysis considerably, as illustrated in the results for the equilibrium solutions presented above.  Therefore we will limit ourselves to a perturbative treatment of the dynamics of the case $n=1$ when $\zeta$ is small.  Inserting the expressions for $a,b$ and $c$ into \eqref{eq:scalarlineardynamics} and collecting like powers of $\zeta$ leads to the equation
\begin{equation}
\frac{d}{dt}\begin{bmatrix}\Ecal \\ \Wcal \\ \Dcal\end{bmatrix}=(A+\zeta B)
\begin{bmatrix}\Ecal \\ \Wcal \\ \Dcal\end{bmatrix},\label{eq:scalarlineardynamicswd}
\end{equation}
where $A$ is as in \eqref{eq:scalarlineardynamics} and 
\begin{equation*}
	B=-\begin{bmatrix}
	2(\beta+\gamma)+1 & \gamma -1 & \beta -1\\
	 \gamma -1& 2(\alpha+\gamma)+1 & \alpha-1 \\
	\beta -1 & \alpha-1  & 2(\alpha+\beta)+1 
\end{bmatrix}.
\end{equation*}
Seeking solutions of the form $ve^{\xi t}$ leads to the eigenvalue problem
\begin{equation*}
	(A+\zeta B)v=\xi v.
\end{equation*}
This can be solved perturbatively for small $\zeta$ \citep{Hinch:1991}. Expressing eigenvectors $v=v^{(0)}+v^{(1)}+O(\zeta^2)$ and eigenvalues $\xi=\xi^{(0)}+\zeta \xi^{(1)}+O(\zeta^2)$ leads to the leading order problem 
\begin{equation}
	Ax^{(0)}=\xi^{(0)} x^{(0)} \label{eq:order1}
\end{equation} 
and the $O(\zeta)$ problem 
\begin{equation}
	(A-\xi^{(0)} I)v^{(1)}=(B-\xi^{(1)} I)v^{(0)}. \label{eq:orderzeta}
\end{equation}
The leading order problem \eqref{eq:order1} is precisely the no-weight-decay case described above, with eigenvalues $\xi^{(0)}_1=-\rho_1=-3,\xi_2^{(0)}=0,\xi_3^{(0)}=0$ and eigenvectors
\begin{equation*}
v^{(0)}_1=\frac{1}{\sqrt{3}}\begin{bmatrix} 1\\ 1\\ 1 \end{bmatrix}, \qquad v^{(0)}_2=\frac{1}{\sqrt{2}}\begin{bmatrix} -1\\ 0\\ 1 \end{bmatrix}, \qquad v^{(0)}_3=\frac{1}{\sqrt{2}}\begin{bmatrix} -1\\ 1\\ 0 \end{bmatrix},
\end{equation*}
The $O(\zeta)$ problem is an inhomogeneous linear system with a singular left-hand side. For a given eigenvalue-eigenvector pair $\xi^{(0)}_i,v^{(0)}_i$, this will only have solutions if the right-hand side lies in the range of $(A-\xi^{(0)}_i I)$, or equivalently is orthogonal to the nullspace of $(A-\xi^{(0)}_i I)^T$. Since $(A-\xi^{(0)}_i I)$ is symmetric, for eigenvalue $\xi^{(0)}=\xi^{(0)}_i,$ the nullspace of $(A-\xi^{(0)}_i I)$ is spanned by $v^{(0)}_i$, and solutions exist if $\left(v^{(0)}_i\right)^T(B-\xi^{(1)} I)v^{(0)}_i=0$. The $O(\zeta)$ correction $\xi^{(1)}_i$ to the $i$th eigenvalue is determined by solving this equation:
\begin{equation}
	\xi^{(1)}_i=\frac{\left(v^{(0)}_i\right)^T B v^{(0)}_i}{\left(v^{(0)}_i\right)^T v^{(0)}_i}.
\end{equation}
Evaluating this yields $\xi^{(1)}_1=3,\xi^{(1)}_2=\xi^{(1)}_3=-1$ (for any choice of $\alpha, \beta,\gamma$ that satisfies $\alpha+\beta+\gamma+1=0$), so with an error of $O(\zeta^2)$ we have
\begin{equation}
	\xi_1=-3+3\zeta, \quad \xi_2=\xi_3=-\zeta.
\end{equation}
Addition of weight decay has a very small detrimental effect on the collective convergence rate $-\xi_1$, but more importantly converts the eigenvalues at zero to negative eigenvalues, leading to decay toward the equilibrium in all directions -- the equilibrium solution becomes asymptotically stable.


\bibliographystyle{unsrtnat}
\bibliography{dim.bib,MDGrefs}

\end{document}